\documentclass[twoside]{article}

\usepackage[accepted]{aistats2026}

\usepackage[round]{natbib}
\renewcommand{\bibname}{References}
\renewcommand{\bibsection}{\subsubsection*{\bibname}}

\usepackage{pdgtools}

   \usepackage[nobox,strip]{restatelinks}

    \usepackage[utf8]{inputenc} %
    \usepackage[T1]{fontenc}    %
    \usepackage{hyperref}       %
    \usepackage{url}            %
    \usepackage{booktabs}       %
    \usepackage{amsfonts}       %
    \usepackage{nicefrac}       %
    \usepackage{microtype}      %
    \usepackage{xcolor}         %
    \usepackage{subcaption}
    \usepackage{caption}
    \let\cite\citep

    \usepackage{wrapfig}
    \usepackage{amsmath,amssymb,mathtools}
    \usepackage{algorithm}
    \usepackage[noend]{algorithmic}
    \usepackage{enumitem}
    \usepackage{bbm}
    \usepackage[capitalize,noabbrev,nameinlink]{cleveref}
    \AddToHook{cmd/appendix/before}{\crefalias{section}{appendix}}

    \usepackage{amsthm,thmtools}
    \theoremstyle{plain}

    \theoremstyle{definition}
    \newtheorem{definition}{Definition}
    
    \theoremstyle{remark}

   \crefname{prop}{Proposition}{Propositions}

    \relax %
       
       \makeatletter
       \setbox0\hbox{$\xdef\scriptratio{\strip@pt\dimexpr
       \numexpr(\sf@size*65536)/\f@size sp}$}
       \newcommand{\scriptveryshortarrow}[1][3pt]{\mathrel{%
       \vcenter{\hbox{\rule[0\fontdimen8\scriptfont3]
       {\scriptratio\dimexpr#1\relax}{\fontdimen8\scriptfont3}}}%
       \mkern-4mu\hbox{\let\f@size\sf@size\usefont{U}{lasy}{m}{n}\symbol{41}}}}
       \makeatother
       \newsavebox{\stosbox}
       \sbox{\stosbox}{\begin{tikzpicture}\draw[->] (0,0) to (0.15,0);\end{tikzpicture}}
       \newcommand\sto{\usebox\stosbox}
       \newsavebox{\stosboxgreen}
       \sbox{\stosboxgreen}{\begin{tikzpicture}\draw[->,green!70!black] (0,0) to (0.15,0);\end{tikzpicture}}
       \newcommand\stogreen{\usebox\stosboxgreen}
       \newsavebox{\stosboxblue}
       \sbox{\stosboxblue}{\begin{tikzpicture}\draw[->,blue] (0,0) to (0.15,0);\end{tikzpicture}}
       \newcommand\stoblue{\usebox\stosboxblue}

     \let\H\relax
    \DeclareMathOperator{\H}{\mathrm{H}} %
    \DeclareMathOperator*{\Ex}{\mathbb{E}} %
    \DeclarePairedDelimiterX{\infdivx}[2]{(}{)}{#1\;\delimsize\|\;#2}
    \newcommand{\thickD}{I\mkern-8muD}
    \newcommand{\kldiv}{\thickD\infdivx}
    \newcommand\mat[1]{\mathbf{#1}}

    \newcommand\numberthis{\addtocounter{equation}{1}\tag{\theequation}}

    \tikzset{factor/.style={draw,minimum width=1.5em, minimum height=1.5em,fill=gray!50!black,text=white,font=\mathversion{bold}\bfseries}}

    \newcommand\btheta{\boldsymbol\theta}

    \newcommand\MThetadense{\dg{M}(\mskip-1mu\Theta\mskip-1mu)}

    \newcommand\Msg{\dg{M\mskip-1mus\mskip-2mu g}}

    \hypersetup{colorlinks=true, linkcolor=blue!75!black, urlcolor=magenta, citecolor=green!50!black}
    \newcommand\vfull[1]{}
    \DeclareMathOperator*{\argmin}{arg\,min} %
    \DeclareMathOperator*{\argmax}{arg\,max} %

    \usepackage[linecolor=black!20,textsize=tiny]{todonotes}

\newcommand\eg{\emph{e.g.}}
\newcommand\ie{\emph{i.e.}}

\begin{document}

\runningauthor{Richardson, Samiei, Shakerinava, Viviano, El Kabid, Parviz, Bengio}

\twocolumn[
\aistatstitle{Local Inconsistency Resolution: The Interplay between Attention and Control in Probabilistic Models}
\aistatsauthor{
    Oliver E. Richardson $^{1,4,5,\dagger}$ \And
    Mandana Samiei $^{2,5}$ \And
    Mehran Shakerinava $^{2,5}$ 
}
\aistatsauthor{%
    ~~~Joseph D. Viviano $^{1,5}$   \And
    ~~~~Abdessamad El Kabid $^{3,5}$ \And
    Ali Parviz $^{5}$ \And
    Yoshua Bengio $^{1,4,5}$
    }
\aistatsaddress{
    \vspace{-1ex}\\
    $^\dagger$Corresponding author: \texttt{oliver.richardson@umontreal.ca}\\[1.5ex]
    $^{1}$Universit\'e de Montr\'eal \hspace{3em}
    $^{2}$School of Computer Science, McGill University  \hspace{3em}
    \\
    $^{3}$Institut Polytechnique de Paris     \hspace{3em}
    $^{4}$LawZero
    \hspace{3em}
    $^{5}$Mila - Quebec Artificial Intelligence Institute
}
]

\begin{abstract}
We present a generic algorithm for learning and approximate inference with an intuitive epistemic interpretation: iteratively focus on a subset of the model and resolve inconsistencies using the parameters under control. This framework, which we call Local Inconsistency Resolution (LIR) is built upon Probabilistic Dependency Graphs (PDGs), which provide a flexible representational foundation capable of capturing inconsistent beliefs. We show how LIR unifies and generalizes a wide variety of important algorithms in the literature, including the Expectation-Maximization (EM) algorithm, belief propagation, adversarial training, GANs, and GFlowNets.
In the last case, LIR actually suggests a more natural loss, which we demonstrate improves GFlowNet convergence.
Each method can be recovered as a specific instance of LIR by choosing a procedure to direct focus (attention and control). We implement this algorithm for discrete PDGs and study its properties on synthetically generated PDGs, comparing its behavior to the global optimization semantics of the full PDG.
\end{abstract}

\section{Introduction}
What causes one to change their mind, to learn, act, and draw inferences? 
From a neurocognitive perspective, the mind operates by resolving internal inconsistencies, whether they arise from conflicting new information or from new awareness of contradictions among prior beliefs.
Theories of cognitive dissonance \cite{festinger1962cognitive} famously explain much of human behavior as targeted adjustments to dispel the discomfort of psychological inconsistency. 
Predictive coding theories \cite{rao1999predictive,friston2005theory} posit that discrepancies between expected and actual sensory inputs---known as prediction errors---are treated as inconsistency signals that drive learning via local adjustments to neural representation.
Inspired by the deeper underlying neurocognitive principle,  we introduce a computational framework that formalizes the process of \emph{local inconsistency resolution} (LIR) in probabilistic models.

The locality---which manifests in two ways---is essential for those of us with computational limitations. 
First, inconsistency can be difficult to detect, and 
   impossible to address constructively
    unless it falls within our field of \emph{attention}. 
Second, we seldom have uniform \emph{control} over our epistemic state;
   we are generally reluctant to revise our direct observations or bedrock principles, forcing us instead to adjust other more pliable beliefs.
So in practice, we resolve inconsistencies \emph{locally}---by looking at only a small part of the model, and modifying 
only a fraction of its parameters.
This process isn't guaranteed to succeed: some inconsistencies simply cannot be seen or addressed with a narrow focus, and fixing one inconsistency can easily create others out of view.
Nevertheless, we will show that LIR provides a powerful recipe for learning and (approximate) inference and demonstrate that many foundational techniques in the literature arise naturally as instances of it.

Our approach leans heavily on the theory of
Probabilistic Dependency Graphs (PDGs), which are highly flexible graphical
   models capable of representing arbitrary (even inconsistent)
   probabilistic information, weighted by confidence \cite{pdg-aaai}.
There is a natural way
   to measure how inconsistent a PDG is,
   and many standard loss functions
   can be viewed as measuring the inconsistency of a PDG that
   describes the appropriate situation \cite{one-true-loss}.
LIR operationalizes training models as
   the process of adjusting parameters to resolve this inconsistency.

Computing a PDG's degree of inconsistency is easy in some cases, 
   but it is often intractable 
   and is NP-hard in general
   (\eg, the log number of solutions to a SAT problem and the log evidence
   of a latent variable model can both be represented as PDG inconsistencies). 
The problem is equivalent to inference in PDGs, 
   which, like for other graphical models, 
   can be achieved in polynomial time under the assumption of bounded tree-width \cite{pdg-infer}.   
Variational inference \citep{blei2017variational,kingma2013autoencoding,jordan1999introduction} can be understood as
   the practice of adopting extra beliefs to
   get an over\-approximation of inconsistency that is easier to calculate \cite{one-true-loss}.
Our approach also enables the opposite:
   focusing on small parts of the graph at a time to
   address tractable under\-approximations of the global inconsistency.

\section{Preliminaries and Parametric PDGs}
\textbf{Variables and Probability.}
We write $\V\! X$ for the set of values that a (random) variable $X$ can take on,
and $\Delta \V \! X $ for the set of distributions over $\V\!X$.
A conditional probability distribution (cpd) is a map
$p(Y|X) : \V \!X \to \Delta\mskip-1mu \V \mskip-1mu Y$.
If $\X = \{X_1, X_2, \ldots\}_{i \in I}$ is an (indexed) set of variables, we regard $\X$ itself as a variable that can take on joint settings $\V\!\X = \prod_{i\in I} \V \! X_i$.
If $\mu \in \Delta \V\!\X$ is a joint distribution and $Y \subseteq \X$, we write $\mu(Y)$ for the marginal of $\mu$ on the variables $Y$. 

   A \emph{directed hypergraph}
   $(N, \mathcal A)$ is a set of nodes $N$ and a set of arcs $\mathcal A$,
   each $a \in \mathcal A$ of which
   is associated with
   a set $\Src a \subseteq N$ of source nodes,
   and $\Tgt a \subseteq N$ target nodes.
   We also write $\ed {\scriptstyle a}{S}{T} \in \Ar$ to specify an
   arc $a$ together with its sources $S = \Src a$ and targets $T = \Tgt a$.

\textbf{Geometry.}
For our purposes, a \emph{pointed parameter space} $\Theta$ is a convex subset of $\mathbb R^n$ for some $n\ge0$
   (that may differ between parameter spaces) with a distinguished default value $\theta_0 \in \Theta$.
A \emph{vector field} over $\Theta$ is a differentiable
   map $X$ assigning to each $\theta \in \Theta$ a vector $X_\theta \in \mathbb R^n$.
The \emph{gradient} of a twice differentiable map $f : \Theta \to \mathbb R$,
   which we write $\nabla_\Theta f(\Theta)$, is a vector field.
Given a vector field  $X$ and an initial point $\theta_0 \in \Theta$, there is a unique trajectory $y(t)$ that solves the ordinary differential equation (ODE)
$\{\frac{\mathrm d }{\mathrm d t}y(t) = X_{y(t)}$,  $y(0) = \theta_0\}$.
To refer to that solution compactly, we adopt the notation $\exp_{\theta_0}( X ) := y(1)$.
Although $\exp$ may appear to give us access only to $y(1)$,
    it is easily verified
    that $\exp_{\theta_0}(t X) = y(t)$ for all $t \ge 0$.
Putting all the pieces together: the map $t\mapsto \exp_\theta(t \nabla_\Theta f(\Theta))$ is the smooth path beginning at $\theta$ that follows the gradient of $f$. It is known as
\emph{gradient flow}.

\vfull{
Given a manifold $\Theta$ and a differentiable map $P : \Theta \to \Delta \V\! X$,
    the Fisher Information Matrix
$\mathcal I(\theta)
$
at each $\theta \in \Theta$
gives rise to a Riemannian metric;
thus the mere fact that $\Theta$ parameterizes a family of
probability distributions is enough to make it a Riemannian manifold.
Moreover, $\mathcal I(\theta)$ is particularly natural in a probabilistic context;
   up to a multiplicative constant, it is the \emph{only} such metric on $\Theta$ that is invariant under sufficient statistics, \cite{chentsov}. \cite{infogoem}
}

\textbf{Probabilistic Dependency Graphs.}
A PDG is a
   directed
   (hyper)graph
   whose arcs carry
   probabilistic and causal information, weighted by confidence \cite{pdg-aaai}.
We define an unweighted (but for our purposes, equivalent) variant whose explicit parametric nature will prove useful.

\begin{definition}
   An \emph{unweighted parametric PDG}
   $
   \dg M(\Theta)$
   $=
   (\X\mskip-2mu, \Ar, \Theta{=} \{\Theta_a \}_{a \in \Ar}, \mathbb P {=}\{ \p_a \}_{a \in \Ar} )$
   is a directed hypergraph
       $(\X\mskip-2mu, \Ar)$
   whose nodes $\X$ are
   variables,
   and whose arcs
   $a \in \Ar$
   are each associated with
       a parameter space $\Theta_a$ 
       and
       a map
       $\p_a : \Theta_a \times \V\Src a \to \Delta \V \Tgt a$
       that gives a cpd
       $\p_{a}(\Tgt a | \Src a;\theta)$
       over $a$'s target variables given values of its sources and a parameter setting $\theta \in \Theta_a$.

   \qedhere
\end{definition}

An unweighted PDG is the object obtained by fixing the parameters:
a joint setting $\btheta = (\theta_a)_{a \in \Ar}$ yields a concrete unweighted PDG
$\dg M = \dg M(\btheta)$---which, at a syntactic level, is just a collection of cpds. 
When constructing PDGs from known cpds (\eg, $p(Y|X)$), we use the mathematical symbol representing that cpd (\eg, $p$) for both the arc $a$ and its associated cpd $\p_a$. 
We typically depict (parametric) PDGs in graphical notation, specifying a cpd 
\vspace{-1.7ex}
\begin{center}
$p_\theta(Y|X,Z)$\;as\;
\begin{tikzpicture}[center base]
   \node[dpadinline] (Y) at (0.95,0) {$Y$};
   \node[dpadinline] (X) at (-0.1,-0.25) {$X$};
   \node[dpadinline] (Z) at (-0.1, 0.25) {$Z$};
   \coordinate (ctr) at (0.38,0);
   \cmergearr[arr2] {X.0}{Z.0}Y{ctr}
   \node[above right=-1pt and -3pt of ctr] {$p$};
\end{tikzpicture}
\;and\; $q_\theta(A,B)$ \;as\;
\begin{tikzpicture}[center base]
   \node[dpadinline] (A) at (1,0.25) {$A$};
   \node[dpadinline] (B) at (1,-0.25) {$B$};
   \coordinate (tip) at (0.15, 0);
   \coordinate (center) at (0.45,0);
   \cunmergearr[arr1] {tip}{A.180}{B.180}{center}
   \node[above left=0pt and 0pt of center,inner sep=2pt]{$q$};
\end{tikzpicture}\,.
\end{center}
\vspace{-1.7ex}

A PDG can be faithfully viewed as the special case of a parametric PDG in which each parameter space 
$\Theta_a = \{ \theta_0 \}$ is a singleton containing only the default value.
Conversely, a parametric PDG may be viewed as a PDG by adding each $\Theta_a$ as an explicit variable.

\textbf{PDG Semantics and Inconsistency.}
The power of PDGs comes from their semantics, which 
sew
their (possibly inconsistent) 
cpds and confidences together to describe features of a joint distribution $\mu(\X)$.

A PDG contains two kinds of information: qualitative structure, encoding
the types of causal mechanisms (the hypergraph $\Ar$),
and observational data
\unskip, in the form of specific probabilistic beliefs
    (the cpds $\mathbb P$).
In the standard presentation \citep{oli-dissertation}, a PDG also comes with weight vectors  $\balpha,\bbeta \in \smash{\overline{\mathbb R}}^\Ar$ over the arcs, encoding confidence in information of each type: $\alpha_a$ can be thought of as the number of independent causal mechanisms by which $\Src a$ determines $\Tgt a$ \cite{QIM}, while $\beta_a$ can be thought of as the effective number of independent reports endorsing that the probability of $\Tgt a$ given $\Src a$ is $\p_a$.
Corresponding to these two types of information, PDG semantics provide a way of scoring compatibility of a joint distribution $\mu(\X)$ with information of each type
\unskip.

With respect to a PDG $\dg M$,
the \emph{observational incompatibility}
of a joint probability measure
$\mu \in \Delta \V\!\X$ is given by
a weighted sum of relative entropy (KL) terms
\begin{equation}
\OInc_{\dg M}(\mu) :=
   \!
   \sum_{\smash{\ed aST \mathrlap{\,\in \Ar}}} \subafalse
   \beta_a\, \kldiv[\Big]{\mu(\Tgt a,\Src a)}{\p_a(\Tgt a | \Src a) \mu(\Src a)}
       ,
       \label{eq:oinc}
\end{equation}
where $\kldiv{p}{q} = \Ex_{x\sim p}[\log \frac{p(x)}{q(x)}]$.
This quantity can be thought of as the excess cost of using codes
optimized for our beliefs $\mathbb P$ weighted by the confidence $\bbeta$ we have in them, when in fact $\X \sim \mu$.
If $\bbeta > \mat 0$, then $\OInc_{\dg M}(\mu) = 0$ if and only if 
$\mu$ satisfies the constraints imposed by every cpd of $\mathbb P$.
Intuitively, the cpds of $\mathbb P$ are inconsistent when constraints are not simultaneously satisfiable,
in which case the \emph{observational inconsistency} $\aar{\dg M}_0 := \inf_\mu \OInc_{\dg M}(\mu)$ 
is an important measure of the magnitude of the unavoidable internal conflict between the probabilistic data in $\dg M$.

Modern machine learning usually prizes observational data above all else---in particular, above structural information such as causal influence or qualitative independencies. For this reason, the scoring function $\OInc$ and the corresponding observational inconsistency $\aar{\dg M}_0$ will suffice for most of our examples. 
That said,
we can also score $\mu$ by its incompatibility
with the structural information in the weighted hypergraph $(\Ar,\balpha)$.
This \emph{structural deficiency} ($\SDef$) is given by:%
   \footnote{If the underlying variables are continuous, then we redefine entropy
   as follows. Assume each variable comes with a base measure $\lambda$, like the Lebesgue or counting measure.
   Then define $\H_\mu(X) := \Ex_\mu[ \log \frac{\mathrm d\mu(X)}{\mathrm d \lambda_X}]$, where $\frac{\mathrm d\mu}{\mathrm d\lambda}$ is the Radon-Nikodym derivative of $\mu(X)$ with respect to $\lambda_X$.}
\begin{align*}
   \SDef_{\!\Ar,\balpha}(\mu) &:=
       \numberthis
        \label{eq:sdef}
       - \H(\X) + \sum_{a \in \Ar} \alpha_a \H(\Tgt a \mid \Src a)
   ,
\end{align*}
and, roughly speaking, measures $\mu$'s failure to arise as a result of
   an independent causal mechanism along each hyperarc \cite{QIM}.
If $\Ar$ is a qualitative Bayesian Network and $\balpha = \mat 1$, for instance,
   then $\SDef_{\!\Ar,\balpha}(\mu) \ge 0$ with equality
   iff $\mu$ has the conditional independencies of $\Ar$.
With confidence $\gamma \ge 0$ in the structural information overall,
the $\gamma$-\emph{inconsistency} of $\dg M$ is the smallest possible overall incompatibility of any distribution with $\dg M$, and denoted
\begin{equation}\label{eq:inconsistency}
   \aar[\big]{\dg M}_\gamma := {\inf_{\mu}\,  \Big( 
       \OInc_{\dg M}(\mu)
        +  \gamma \, \SDef_{\!(\Ar,\balpha)}(\mu) \Big).}
\end{equation}
\citet{one-true-loss} argues that this inconsistency measure
   \eqref{eq:inconsistency}
   is a ``universal'' loss function, largely by showing how it
   specializes to standard loss functions across a wide breadth of contexts.
It follows that, at least at an abstract level,
   much of machine learning can be viewed as inconsistency resolution.
We now take this idea a few steps further, by operationalizing the
   resolution process
   and allowing it to be done ``locally''.

\section{The Local Inconsistency Resolution (LIR) Algorithm}

There are two distinct senses in which inconsistency resolution can
   be \emph{local}: we can restrict what we can see, or what we can do about it.
Correspondingly, there are two ``focus'' knobs for our algorithm:
   one restricts our \textbf{attention} to the inconsistency of a subset of arcs $A \subseteq \Ar$,
   and the other restricts our \textbf{control} to only the parameters of a subset
   $C
    \subseteq \Ar$
    of arcs
     as we resolve that inconsistency.
The former makes for an underestimate of the inconsistency that is easier to calculate, while
the latter
makes for an easier optimization problem.
These restrictions are not just cheap approximations, though:
   they are also appropriate modeling assumptions for
   actors that cannot see and control everything at once.

Attention and control need not be black and white.
A more general approach would be to choose
   an \emph{attention mask} $\varphi \in \smash{\overline{\mathbb R}}^{\Ar}$ and
   a \emph{control mask} $\chi \in [0,\infty]^{\Ar}$.
Large $\varphi(a)$ makes $a$ salient while $\varphi(a) \!=\! 0$ keeps it out of mind;
similarly, large $\chi(a)$ gives significant freedom to change $a$'s parameters,
small $\chi(a)$ affords only minor adjustments, and $\chi(a) \!=\! 0$
   prevents change altogether.
We often say ``full control'' to mean $\chi(a) =\infty$, in which case we move $\theta_a$ to a fixed point.

In full generality, we refine the types of $\varphi$ and $\chi$ one step further by breaking the control and attention in each arcs into natural subcomponents. 
The behavior of the attention mask $\varphi$ described in the previous paragraph matches the role played by $\balpha$ and $\bbeta$ in PDG semantics---which is why we needed only to define unweighted PDGs as our basic objects, but require weighted semantics for them.
The parameter $\gamma$ also describes a form of attention, to the overall qualitative information in the network. 
Thus, we can reinterpret all three classical weight parameters as components of the attention mask $\varphi = (\balpha,\bbeta,\gamma)$, a vector of dimension $2|\Ar|+1$.
To simplify notation, we write $\aar{\varphi \odot \dg M} := \aar{\dg M,\balpha,\bbeta}_\gamma$ to denote the $\gamma$-inconsistency of the weighted PDG $(\dg M, \balpha,\bbeta)$. 
We will typically set $\gamma=0$, rendering $\balpha$ irrelevant (except in \cref{sec:factor-graph} where $\gamma=1$ and $\balpha=\bbeta$), and also assume
 $\bbeta = \balpha = \mat 1$ unless otherwise specified. 
We also allow $\chi(a)$ to be a vector of dimension $\dim \Theta_a$, so overall $\chi$ is a vector of dimension $\sum_{a \in \Ar} \dim \Theta_a$.

\begin{algorithm}
   \caption{Local Inconsistency Resolution (LIR)}
   \label{algo:LIR}
   \begin{algorithmic}
       \STATE \textbf{Require:} procedure {\sc Refocus}
       \STATE \textbf{Input:}
           knowledge base $\MThetadense$,

       \STATE Initialize $\btheta^{(0)} \gets 
           0$;

       \FOR{$t = 0, 1, 2, \ldots$}
           \STATE $\varphi, \chi
               \gets \textsc{Refocus}()$;
           \STATE $\btheta^{(t+1)} \!\gets \exp_{\btheta^{(\mskip-1mut\mskip-1mu)}}\!
               \Big\{\! {-} \chi \odot \nabla_{\!\btheta}
               \aar[\Big]{ \varphi\odot \dg M(\btheta) } \Big\}$;
       \ENDFOR
   \end{algorithmic}
\end{algorithm}

Formalized in \cref{algo:LIR}, LIR
   is a heuristic algorithm that adjusts the parameters of a parametric PDG $\MThetadense$
   as to locally resolve inconsistencies as follows.
Start with a belief state, in the form of a parametric PDG $\dg M(\Theta)$, and initialize the parameters to their default values.
In each iteration, choose \emph{focus} consisting of an attention mask $\varphi = (\balpha,\bbeta,\gamma)$ and a control mask $\chi$.
Calculate
$\aar{\varphi\odot\MThetadense}_\gamma$, the inconsistency of
   the combined context and memory, weighted by attention.
In many cases (\eg, when $\bbeta \ge \gamma\balpha$) these problems can be solved with techniques in conic optimization \citep{pdg-infer},
   but this may be intractable unless the attention is narrow or one can find a formula for it in closed form.
Finally, resolve this local inconsistency
   by updating mutable memory via (an approximation to) gradient flow,
   changing the parameters associated with $a$ in proportion to the degree of control $\chi(a)$.

The ODE described in the final line may be approximated with an inner loop running an iterative
   gradient-based optimization algorithm.
Alternatively, if \textsc{Refocus} produces small $\chi$,
   then it is well-approximated by a single gradient descent step of size $\chi$.
At the other extreme, if $\chi$ is infinite in every component, then
we typically expect
the final line to reduce to
\vspace{-0.5ex}
   \begin{equation}
       \btheta^{(t+1)} \gets 
        \smash{%
        \arg\min_{\btheta} \,
           \aar[\big]{ \varphi\odot \dg M(\btheta) }\,,
        }
           \label{eqn:solve-opt}
   \end{equation}
at least in many cases of interest. 
For example:

\begin{linked}{prop}{logccave}
   If $\dg M(\Theta)$ is an unweighted parametric PDG whose parameterizations
   $\mathbb P$ are either constant or unconditional and log-concave and
   $\bbeta \ge \gamma\balpha$,
   the map $\btheta \mapsto \aar{\dg M(\btheta),\balpha,\bbeta}_\gamma$ is convex.%
   \onlyfirsttime{\footnote{All proofs can be found in the appendix.}}
\end{linked}
Surprisingly, inconsistency is not always convex in the parameters of \emph{conditional} probability distributions---yet we conjecture that \eqref{eqn:solve-opt} holds nevertheless. Determining whether or not $\aar{\dg M + p(Y|X)}$ is quasi-convex in $p$ remains a key open question in the theory of PDGs.

In order to run \cref{algo:LIR},
we must select the procedure 
\textsc{Refocus} that provides attention and control masks.
We focus primarily on the case where \textsc{Refocus}
   chooses non-deterministically
   from a fixed finite set of attention-control mask pairs $\mathbf{F}
   = \{ (\varphi_i,\chi_i) \}_{i=1}^n
$, which we call \emph{foci}.
Even so, 
\cref{algo:LIR} is quite abstract. 
To give intuition for it, next we give a sample of  important algorithms that are instances of LIR.

\section{Unifying Important Algorithms as Instances of LIR}

\subsection{The Classification Setting}
    \label{sec:classification}

\begin{figure}
   \centering
   \tikzset{atkv/.style={green!70!black},defv/.style={blue}}
   \begin{tikzpicture}[center base]
       \node[dpad1] (X) at (0,0) {$X$};
       \node[dpad1] (Y) at (1.3,0) {$Y$};

       \draw[arr1,-,shorten >=3pt,transform canvas={yshift=-1pt}, defv,dashed,ultra thick] (X) to (Y);
       \draw[arr1,-,shorten >=3pt,transform canvas={yshift=0.75pt}, atkv] (X) to (Y);
       \draw[arr1] (X) to node[above] {$p$} (Y);
       \coordinate (xend) at ($(X)+(-1.1,0.5)$);
       \draw[arr1,-,shorten <=5pt,transform canvas={yshift=0.75pt}, atkv] (X) to (xend);
       \draw[arr1, <-] (X) to
            node[above,pos=0.65]{$x$}
            (xend);
       \coordinate (x'end) at ($(X)+(-1.1,-0.5)$);
       \draw[arr1,-,ultra thick,shorten <=4pt,transform canvas={yshift= 1pt}, atkv, dashed] (X) to (x'end);
       \draw[arr1,-,shorten <=3pt,transform canvas={yshift= -0.75pt}, defv] (X) to +(x'end);
       \draw[arr1, <<-] (X) to node[above,pos=0.65]{$x'$} (x'end);
       \coordinate (yend) at ($(Y)+(1,0.5)$);
       \coordinate (y'end) at ($(Y)+(1,-0.5)$);
       \draw[arr1,-,shorten <=6pt,transform canvas={yshift=-0.75pt}, defv] (Y) to (yend);
       \draw[arr1, <<-] (Y) to node[above,pos=0.65]{$y$} (yend);
       \draw[arr1,-,shorten <=6pt,transform canvas={yshift=0.75pt}, atkv] (Y) to (y'end);
       \draw[arr1, <<-] (Y) to node[above,pos=0.65]{$y'$} (y'end);
   \end{tikzpicture}
   $\cong$
   \begin{tikzpicture}[center base]
       \def\spray{.24}
       \node[dpad1] (X) at (0,-0.6) {$X$};
       \node[dpad1] (Y) at (1.4,0) {$Y$};
       \node[dpad1] (T) at (0, 0.6){$\Theta$};
       \mergearr[arr1]XTY
       \node[above right=0pt and 0pt of center-XTY]{$p$};
       \coordinate (tend) at ($(T)+(-1.1,\spray)$);
       \draw[arr1,-,ultra thick,shorten <=7pt,transform canvas={yshift= -1.0pt}, defv, dashed] (T.160) to (tend);
       \draw[arr1,-,shorten <=3pt,transform canvas={yshift= 0.7pt}, atkv] (T.160) to (tend);
       \draw[arr1, <<-] (T.160) to node[left,pos=0.9]{$\theta$} (tend);
       \coordinate (nend) at ($(T)+(-1.1,-\spray)$);
       \draw[arr1, <-, defv] (T.-160) to
           node[below,inner sep=2px,pos=0.7]{\scalebox{0.8}{$\mathcal N(0,1)$}} (nend);
       \coordinate (xend) at ($(X)+(-1.1,\spray)$);
       \draw[arr1, <-, atkv] (X.160) to
           node[above,inner sep=1px,pos=0.7]{\scalebox{0.8}{$\mathcal N(x,1)$}} (xend);
       \coordinate (x'end) at ($(X)+(-1.1,-\spray)$);
       \draw[arr1,-,ultra thick,shorten <=6pt,transform canvas={yshift=1.0pt}, atkv, dashed] (X.-160) to (x'end);
       \draw[arr1,-,shorten <=3pt,transform canvas={yshift= -0.7pt}, defv] (X.-160) to (x'end);
       \draw[arr1, <<-] (X.-160) to node[left,pos=0.9]{$x'$} (x'end);
       \coordinate (yend) at ($(Y)+(1,\spray)$);
       \coordinate (y'end) at ($(Y)+(1,-\spray)$);
       \draw[arr1, <<-,defv] (Y.35) to node[above,pos=0.65]{$y$} (yend);
       \draw[arr1, <<-,atkv] (Y.-35) to node[below,pos=0.65]{$y'$} (y'end);
   \end{tikzpicture}
   \caption{\small 
       Illustrations of adversarial training as LIR.
       Foci are in green and blue; dashes indicate control.
       Right: explicit parameter variable $\Theta$ and exact symmetry.
   }%
   \label{fig:adv}
\end{figure}
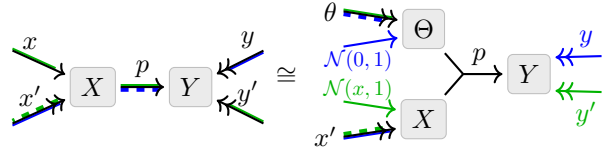

Consider a parametric classifier $p_\theta(Y|X)$, perhaps
   arising from a neural network whose final layer is a softmax,
   and suppose we also have a labeled example $(x,y)$. 
Let's capture this situation with a parametric PDG.
Since $\V Y$ is a discrete label space space, we regard labels as unconditional distributions over $Y$, viewing ``hard'' labels such as our $y \in \V Y$ as vertices of this simplex, while also allowing for label smoothing \citep[see][for an overview]{muller2019does}.
Doing the same for $x$ may be intractable, but fortunately is not necessary; in the typical case where the space of inputs $\V\!X$ is itself a manifold (\eg, color images in $[0,1]^{W\times H\times 3}$), we can regard a value $x \in \V\!X$ as parameterizing a deterministic unconditional probability over $X$.
To communicate this deterministic parameterization visually, we use a double-headed arrow, writing
$\smash{
\tikz[center base]{\node[dpadinline](X) {$X$}; \draw[arr1, <<-](X) to node[fill=white,inner sep=1pt,pos=0.65]{$x$} +(-1.1,0);}
}$.
Combining the parametric classifier $p_\theta(Y|X)$ with the sample $(x,y)$,
   we get a parametric PDG $\dg M(x,y,\theta) := $
\vspace{-2ex}
\begin{equation}
   \;
   \begin{tikzpicture}[baseline=-0.6ex]
       \node[dpadinline] (X) at (0,0) {$X$};
       \node[dpadinline] (Y) at (1.1,0) {$Y$};
       \draw[arr1, <<-] (X) to node[above,pos=0.65]{$x$} +(-0.85,0);
       \draw[arr1, <-] (Y) to node[above,pos=0.65,inner sep=2pt]{$y$} +(0.75,0);
       \draw[arr1] (X) to
           node[above] {$p_\theta$} (Y);
   \end{tikzpicture}
   \quad\cong\quad
       \begin{tikzpicture}[center base]
       \def\spray{.24}
       \node[dpad0] (X) at (0,-0.4) {$X$};
       \node[dpad0] (Y) at (1.2,0) {$Y$};
       \node[dpad0] (T) at (0, 0.4){$\Theta$};
       \mergearr[arr1,gray]XTY
       \node[above right=0pt and 0pt of center-XTY]{$p$};
       \draw[arr1, <<-] (X) to node[above,pos=0.65]{$x$} +(-0.85,0);
       \draw[arr1, <<-] (T) to node[above,pos=0.65]{$\theta$} +(-0.85,0);
       \draw[arr1, <-] (Y) to node[above,pos=0.65,inner sep=2pt]{$y$} +(0.75,0);
   \end{tikzpicture}
   \label{pdg:classfication}
\end{equation}
whose observational inconsistency is
$
   - \log p_\theta(y|x)
$, the standard training objective for such a classifier \cite{one-true-loss}.
Each cpd plays major role in this inconsistency.
What happens when we resolve this it
with control over different arcs?

\begin{itemize}[nosep,itemsep=3pt, left=0.5em]
   \item Adjusting $\theta$
   amounts to \textbf{training} the network in the standard way.
       In this case, the value $\chi$ of the control mask corresponds roughly
       to the product of the learning rate and the number of optimization iterations spent on the example $(x,y)$.

   \item Adjusting $y$
   amounts to \textbf{inference}:
   it adjusts $y$ to match distribution $p_\theta(Y|x)$.

   \item Adjusting $x$
   amounts to \textbf{forming an adversarial example}: it makes small changes to the input
   until the (fixed) network gives it the desired label.
\end{itemize}

Some readers may find the last point surprising. 
While adversarial examples \citep{goodfellow2014explaining} are often presented as weird quirk of neural networks,
   they are, together with training and inference, one of the three basic resolutions ways of resolving this simple inconsistency.
The sustained interest of the machine learning community on adversarial examples
   may appear to be a cultural phenomenon,
   but at a mathematical level, it is no accident \citep{shafahi2018adversarial}.
At this level of abstraction, there is no difference between
   the network parameters and inputs.
Making the parameterization of $p$ explicit and adding L2 regularization,
the symmetry
becomes striking (see \cref{fig:adv}, right).

With minor modifications to $\dg M(x,y,\theta)$, we can capture variations of training procedures.

\textbf{Stochastic Gradient Descent (SGD).} \label{sec:SGD}
Take the mutable state to be the classifier $p$ as before.
Define $\textsc{Refocus}$ so that it draws a batch of samples $\{(x_i,y_i)\}_{i=1}^m$, places infinite attention on their empirical distribution $d(X,Y)$ (reflecting high confidence in the data).
If $\eta := \chi(p) \varphi(p)$ is small, then
   LIR reduces to SGD with batch size $m$ and learning rate $\eta$.

\textbf{Adversarial training.}
Suppose we want to slightly alter $x$ to obtain $x'$ that is classified as $y'$ instead of $y$.
Adding $x'$ and $y'$ to $\dg M$ and relaxing $\p_x$ to be a Gaussian centered $x$ rather than a point mass,
we get the PPDG on the left of \cref{fig:adv}.
A LIR iteration whose focus consists of the edges marked in green (with control over the dashed green edge)
   is an adversarial attack with Euclidean distance \cite{biggio2013advattk}.
The blue focus, by contrast, ``patches'' the adversarial example by
   adjusting the model parameters to again classify it correctly.
Thus, LIR that alternates between the two and refreshes $(x,y,y')$ is adversarial training, a standard defense to adversarial attacks \cite{goodfellow2014explaining}.
Thus, it is just as sensible to train the inputs, as the network \cite{FNNS}.

\subsection{The EM Algorithm and VAEs}

Consider a latent variable model $p_\theta(Z,X)$.
Given observations of the form $X{=}x$,
how can we learn model parameters $\theta$ despite the missing data $z$?
The standard answer is
the EM algorithm \citep{dempster1977maximum,mclachlan2008algorithm},
which iteratively improves an estimate of $\theta$ by first taking an 
\textbf Expectation over the missing values $z$ (fixing the current estimate $\theta$), and then performing
\textbf Maximization over $\theta$ to update the parameters;
altogether this amounts to computing:
\[
   \theta^{(t+1)}_{\text{EM}}
           := \arg\max_{\theta}\; \Ex\nolimits_{z\sim p(Z|x,\theta_{\text{EM}}^{(t)})} \big[ \,\log p(x, z | \theta) \,\big].
\]
\vspace{-2.5ex}

\begin{linked}{prop}{LIR-EM}
   \textsc{LIR}$\Bigg(\!
       \begin{tikzpicture}[center base]
           \node[dpad0] (Z) {$Z$};
           \node[dpad0,left=.5 of Z] (X) {$X$};
           \draw[arr2, <<-] (X) --  node[above,pos=0.8]{$ x$} +(-0.9, 0);
           \coordinate (A) at ($ (X)!.5!(Z) + (0,0.8)$);
           \draw[arr1] (A) -- node[left, inner sep=3pt]{$p$} ++(0,-0.35) -- (Z);
           \draw[arr1] (A) -- ++(0,-0.35) -- (X);
           \draw[arr2, <-] (Z) --
               node[above,pos=0.65, inner sep=2pt]{$q$}
               node[below,pos=0.7, inner sep=2pt]{${\color{gray}\scriptscriptstyle(\infty)}$}
               ++(0.9, 0);%
       \end{tikzpicture}
   \!\Bigg)$
   in which {\sc Refocus} re-samples $x$ 
   and alternates between
   full control of $p$ and $q$
   implements EM, in that
   $\theta_{\text{EM}}^{(t)} = \theta_{\text{LIR}}^{(2t)}$%
   .%
   \onlyfirsttime{\footnote{The result can also be readily adapted to an entire dataset by replacing $x$ with a high confidence empirical distribution, or mini-batched as in \cref{sec:SGD}.}}
\end{linked}

This result is closely related to one due to
\citet{neal1998view}, who view it as an intuitive explanation of why the EM algorithm works.  Indeed, it is obvious in this form that every adjustment reduces the overall inconsistency, and that
the M step need not fully solve the maximization problem, but merely make progress.

\textbf{Variational inference as LIR.}
This form of the EM algorithm is closely related to variational inference.
Indeed, the inconsistency of the PDG in \cref{prop:LIR-EM} is equal to the evidence lower bound (ELBO), the optimization target for variational methods.
Furthermore, allowing $q$ to depend on $x$ 
and breaking up $p(X,Z) = p(Z)d(X|Z)$ into two separate arcs yields a PDG containing the data of a variational autoencoder, or VAE \cite{kingma2013autoencoding}, and the resulting inconsistency is the variant of the ELBO used to train VAEs \cite{one-true-loss}.  It follows that LIR where we control the encoder and decoder (but not the prior) and attend to various mini-batches of samples, trains a VAE with stochastic gradient descent. 
\emph{Mean-field} variational methods are ones where the latents $Z = (Z_i)_{i=1}^n$ are real-valued, and the variational family is restricted to those of the form $q(Z) = \prod_{i=1}^n q_i(Z)$.

\subsection{Generative Adversarial Networks}
\def\pdata{p_{\mathrm{data}}}
\def\real{{\mathrm{real}}}
\def\fake{{\mathrm{fake}}}
LIR can also capture the procedure for training Generative Adversarial Networks, or GANs \cite{goodfellow2020generative}.
The goal is to train a generator network $G$ to generate images that cannot be distinguished
   from real ones.
Formally, let $X$ be either a generated image $X_{\fake}\sim G$ or one from a dataset
$X_\real \sim \pdata$, based on the outcome of a fair coin $C$.
A discriminator $D$ then predicts the outcome $C$ of the coin flip from the image $X$.
The only additional ingredient we will need that is not explicit in the original formulation is a belief $e$ that the coin is equally likely heads as tails given $X$---intuitively, representing the goal of the generator and the opposite of the goal of the discriminator.
All of these pieces of probabilistic information can be summarized by the (foci-annotated) PDG below.
\vspace{-1ex}
\[
   \dg M(\Theta) :=
   \begin{tikzpicture}[center base,Dcolor/.style={green!70!black},Gcolor/.style={blue}]
       \node[dpad1](Xfake) at (0,1) {$X_{\fake}$};
       \node[dpad1](Xreal) at (0,0){$X_{\real}$};
       \node[dpad1](X) at (2.1,0){$X$};
       \node[dpad1](C) at (2.4,1.0){$C$};

       \draw[arr1,-,line width=1.3pt,transform canvas={yshift=1pt},dashed,Gcolor, shorten <=5pt] (Xfake) to +(-1.5,0);
       \draw[arr1, <-] (Xfake) to node[above,pos=0.6]{$G$}
           +(-1.5,0);
       \draw[arr1, <-] (Xreal) to node[above,pos=0.6]{$\pdata!$} +(-2,0);
       \draw[arr,-,line width=1.3pt,transform canvas={xshift=1pt}, Dcolor,dashed]
                  (X.-2) to[out=0,in=-20,looseness=2.21] (C.-28);
       \draw[arr] (X.0) to[out=0,in=-20,looseness=2.2] node[right]{$D$} (C.-30);
       \draw[arr1,-,line width=1.1pt,transform canvas={xshift=0.5pt}, Gcolor,shorten >=3pt]
           (X.10) to[out=10,in=-55,looseness=1.6] (C.-50);
       \draw[arr1,-,line width=1.1pt,transform canvas={xshift=-0.8pt}, Dcolor,shorten >=3pt,shorten <=1pt]
           (X.10) to[out=10,in=-55,looseness=1.6] (C.-50);
       \draw[arr1]
           (X.10)
           to[out=10,in=-55,looseness=1.6]
           node[left,pos=0.6,inner sep=4pt]{$e$}
           (C.-50);
       \coordinate (ctr) at (1.2,0.1);
       \draw[arr1,-,shorten >=0pt] (Xfake) to[bend right=8] (ctr);
       \draw[arr1,-,shorten >=0pt] (Xreal) to[bend left=5] (ctr);
       \draw[arr1,-,shorten >=0pt] (C) to[in=150,out=180,looseness=1.5](ctr);
       \draw[arr1,->>,shorten <=0pt] (ctr) to[bend right=10](X);
       \draw[arr1, <-] (C) to node[above,pos=0.6]{\small$50/50!$} +(1.5,0);
\end{tikzpicture}
\]
GAN training is often written as a 2-player game
$
   \min_{G} \max_{D}  \mathcal L^{\text{GAN}}(G,D)
$, where
$\mathcal L^{\text{GAN}}(G,D) = $
\[
   \Ex\nolimits_{x \sim \pdata}
   [\,\log D(x)\,] + \Ex\nolimits_{x' \sim G} [\,\log (1- D(x'))\,].
\]

\begin{figure*}
\centering
    \begin{tikzpicture}[xscale=2.1,yscale=1.3,center base]
       \node[factor] (a) at (0,0) {$a$};
       \node[draw,circle,inner sep=4px] (X) at (1,0) {$X$};
       \node[factor] (b1) at (2,0.5){$b_1$};
       \node[factor] (bm) at (2,-0.5){$b_m$};
       \node[draw,circle,inner sep=2px] (Y1) at (-1,0.5){$Y_1$};
       \node[draw,circle,inner sep=2px] (Yn) at (-1,-0.5){$Y_n$};
       \draw (a) -- (X);
       \draw (b1) -- (X);
       \draw (bm) -- (X);
       \draw (Y1) -- (a);
       \draw (Yn) -- (a);

       \begin{scope}[transform canvas={yshift=2px},green!70!black]
       \draw[arr,arrows={->[harpoon,swap]}]
           (b1) to node[above=-0.5pt,rotate=15]{$m_{b_1 \!\stogreen \mskip-2muX}$\!\!} (X);
       \draw[arr,arrows={->[harpoon,swap]}]
           (bm) to node[above=-0.5pt,rotate=-5,pos=0.3]{$m_{b_m \!\stogreen \mskip-2muX}$\!\!} (X);
       \draw[arr,arrows={->[harpoon,swap]},densely dotted]
           (X) to node[above]{$m_{X \!\stogreen a}$\!} (a);
       \end{scope}

       \begin{scope}[transform canvas={yshift=-2px},blue]
       \draw[arr,arrows={->[harpoon,swap]}]
           (Y1) to node[below=-0.5pt, rotate=-5,pos=0.35]{$m_{Y_{\!1} \!\stoblue a}$\!\!} (a);
       \draw[arr,arrows={->[harpoon,swap]}]
           (Yn) to node[below, rotate=17]{$m_{Y_{\!n} \!\stoblue a}$\!\!} (a);
       \draw[arr,arrows={->[harpoon,swap]},densely dotted]
           (a) to node[below]{\!$m_{a \stoblue \mskip-2muX}$} (X);
       \end{scope}
   \end{tikzpicture}
   $\qquad
   \begin{tikzpicture}[xscale=2,center base]
   \begin{scope}[os1/.style={outer sep=1pt},dpad0/.append style={fill=gray!50!black,text=white,font=\mathversion{bold}}]
       \node[dpad0,os1] (aX) at (0.22,0){\scalebox{0.7}{$X^{a}$}};
       \node[dpadded,os1] (X) at (1,0) {$X$};
       \node[dpad0,os1] (b1X) at (1.8,0.5){\scalebox{0.7}{$X^{b_1}$}};
       \node[dpad0,os1] (bmX) at (1.8,-0.5){\scalebox{0.7}{$X^{b_m}$}};
       \node[dpadded,os1] (Y1) at (-1.1,0.5){$Y_1$};
       \node[dpadded,os1] (Yn) at (-1.1,-0.5){$Y_n$};
       \node[dpad0,os1] (aY1) at (-0.2,0.3){\scalebox{0.7}{$Y_{1}^{a}$}};
       \node[dpad0,os1] (aYn) at (-0.2,-0.3){\scalebox{0.7}{$Y_{n}^{a}$}};
   \end{scope}
   \begin{scope}[every path/.append style={gray!50,double equal sign distance}]
       \draw (aX) to (X);
       \draw (aY1) to (Y1);
       \draw (aYn) to (Yn);
       \draw (b1X) to (X);
       \draw (bmX) to (X);
   \end{scope}

   \begin{scope}[green!70!black]
       \draw[arr0,
           ]
           (1.5,0.8) to[in=20,out=-80] node[left, pos=0.1]{$m_{b_1 \!\stogreen \mskip-2muX}$\!} ([yshift=1px]X.24);
       \draw[arr0,
           ]
           (1.6,0.1) to[in=-23,out=-95] node[right, pos=0.25]{$m_{b_m \!\stogreen \mskip-2muX}$\!} ([yshift=1px]X.-15);
       \draw[arr0,
               dashed]
           (0.6,0.7) to[in=0,out=-90] node[left, pos=0.1]{$m_{X \!\stogreen a}$\!} ([yshift=2px]aX.east);
   \end{scope}
   \begin{scope}[blue]
       \draw[arr0,
           ]
           (-0.85,-0.1) to[in=160,out=110] node[right, pos=0.1]{$m_{Y_1 \!\stoblue a}$\!} ([yshift=-2px]aY1.175);
       \draw[arr0,
           ]
           (-0.8,-0.9) to[in=-157,out=100] node[right, pos=0.1]{$m_{Y_n \!\stoblue a}$\!} ([yshift=-1px]aYn.-166);

       \draw[arr0,
           dashed]
           (0.6,-0.7) to[in=180,out=90] node[right, pos=0.1]{\!$m_{a \stoblue \mskip-2muX}$} ([yshift=-2px]X.west);
   \end{scope}
   \begin{scope}[blue!50!black]
       \coordinate (amerge) at (-70:0.6);
       \draw[arr0,shorten <=0] (amerge) to[out=130,in=-90] (aX);
       \draw[arr0,shorten <=0] (amerge) to[out=130,in=-45] (aY1);
       \draw[arr0,shorten <=0] (amerge) to[out=130,in=-10] (aYn);
       \draw[arr0,-,shorten >=0] (0.3,-0.9) to[in=-50,out=90]
           node[left]{$\phi_a$} (amerge);
   \end{scope}
\end{tikzpicture}
\raisebox{-1em}{
\!\!
$\subseteq\begin{matrix}\Msg\\+
   {\color{blue!50!black}\dg M_\Phi}
   \text{.\!}
\end{matrix}$}
$

\caption{\small
Left: an illustration of the local message-passing updates (\ref{eq:X->a},\ref{eq:a->X}). Right: the PDG $\Msg$ of \cref{sec:factor-graph}.}
\label{fig:mp}
\end{figure*}
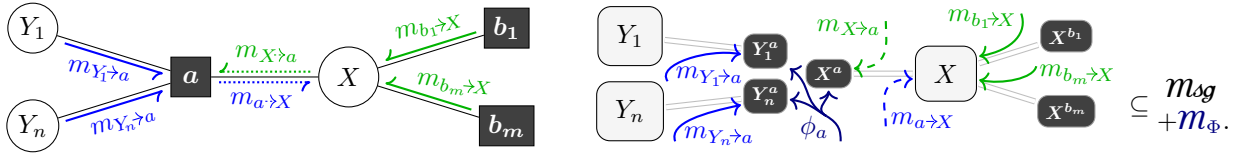

\textbf{The Discriminator's Focus $(\varphi_D, \chi_D)$.}
The discriminator has control over $D$, and attends to
everything but $e$.
That inconsistency of this PDG is what might be called
the discriminator's objective:
the expected KL divergence from $D$ to the optimal discriminator.
If $D$ also disbelieves that any image is equally likely to be fake as real
(by choosing to give negative attention, $\varphi_D(e) = -1$),
then the inconsistency becomes $-\mathcal L^{\text{GAN}}$.

\textbf{The Generator's Focus $(\varphi_G, \chi_G)$.}
The generator has control over $G$.
If it ignores $D$ and attends only to $e$, the inconsistency
is the Jensen-Shannon Divergence between $G$ and $\pdata$.
If the generator also disbelieves the discriminator $D$
(\ie, $ \varphi_G(D) =-1$),
then the inconsistency becomes $+\mathcal L^{\text{GAN}}$.
It follows that:

\begin{linked}{prop}{GAN}
    $\textsc{LIR}(\dg M)$ with foci alternating between $(\varphi_G, \chi_G)$ and $(\varphi_D, \chi_D)$ trains a GAN.
\end{linked}

\vfull{Standard practice is to select $\chi_G(G) \ll \chi_D(D)$.}

\subsection{Message Passing}
   \label{sec:factor-graph}

A \emph{factor graph} over a set of variables $\X$ is a set of factors
$\Phi = \{ \phi_a : \mathbf X_a \to \mathbb R_{\ge 0}\}_{a \in \Ar}$,
where each $\mathbf X_a \subseteq \mathcal X$ is called the \emph{scope} of $a$.
Conversely, for $X \in \X$, let
$\partial X
    := \{a \in \Ar : X \in \mat X_a\}
$ be the set of factors with $X$ in scope.
The factor graph
$\Phi$ specifies a distribution
$\Pr_\Phi(\X) \propto \prod_a \phi_a(\mat X_a)$, and
corresponds to a PDG
$
   \dg M_\Phi = \Big\{ \begin{tikzpicture}[center base]
       \node[dpadinline] (X) {$\mathbf X_a$};
       \draw[arr,<-] (X) to node[above, pos=0.65,inner sep=3pt]{$\propto \smash{\phi_a}$}
           node[below,pos=0.7,inner sep=1.5pt]{\color{gray}\scriptsize$(\alpha,\beta{=}1)$} +(-1.4,0);
   \end{tikzpicture}~\Big\}_{a \in \Ar}
$
that specifies the same distribution $\Pr_\Phi(\X)$ when $\gamma{=}1$.

Sum-product belief propagation (BP) \cite{kschischang2001factor}
   aims to approximate marginals of $\Pr_{\Phi}$
   with local computations: interactions between 
    neighboring factors and variables. 
The state consists ``messages''
$m_{X \sto a}$ and $m_{a \sto X}$
both (unnormalized) distributions over $X$,
for each variable $X$ and factor $a \in \partial X$ adjacent to it.
After initialization, BP repeatedly recomputes:
\begin{align}
   m_{X \!\sto a}(x)
       &:\propto
       \prod_{{b \in \partial X\setminus a}} m_{b\sto\! X} (x)
       \label{eq:X->a}
       \\
   m_{a \sto\mskip-2mu X}(x)
       &:\propto
       ~~\sum_{\mathclap{\mat y \in \V(\mat X_a \setminus X)}}~~ \phi_a(x, \mat y)
       ~\prod_{\mathclap{Y \in \mat X_a \setminus X}}~
               m_{Y \!\sto a} (Y(\mat y)),
       \label{eq:a->X}
\end{align}
\vspace{-1ex}
where $Y(\mat y)$ is the value of $Y$ in the joint setting $\mat y$.

Observe that these calculations are (marginals of) products of factors,
   and thus correspond to inference
       in a ``local'' factor graph.
The usual schematic illustration people draw to depict messages
   moving between variables and factors according to \cref{eq:X->a,eq:a->X}
   is given on the left of \cref{fig:mp}.
While this standard diagram is only a schematic,
   simply writing down the messages as unconditional probability
   distributions, we get a PDG $\Msg$ that can be made to look very similar.
Formally, a variable $X^{a}$ for every pair
   $(X,a)$ with $X \in \mat X_a$
along with edges asserting that $X^{a} = X$,
we obtain the equivalent PDG on the right of \cref{fig:mp}.
Indeed, (\ref{eq:X->a},\ref{eq:a->X}) minimize inconsistency of
   the controlled beliefs in their appropriate contexts
   (illustrated in \cref{fig:mp} (right) and formalized
       in \cref{sec:bp-details}).

At the end, variable marginals $\{b_X\}_{X \in \X}$,
which we regard as another PDG $\dg B$, are computed from the messages according to
$
   b_X(x) \propto \prod_{a \in \partial X} m_{a \sto X}(x)
$.

\begin{linked}{prop}{bp}
   If {\sc Refocus} selects a focus non-deterministically from
   $\{ a\sto\mskip-2mu X, X\! \sto a, X\}_{X \in \X, a \in \partial X}$
   (see illustration above; details in \cref{sec:bp-details}), then
   the possible runs of
   {\sc LIR}$(
       \dg M_\Phi, \Msg
       + \dg B
        )$
   are precisely those of BP for different message schedules.
\end{linked}

The same construction works for general cluster graphs,
   and we suspect it also reproduces the many message passing algorithms
   generated by $\alpha$-divergences \cite{minka2005divergence},
   because of the close relationship those divergences have with PDGs
   \cite[\S5]{one-true-loss}.

\subsection{Transformer Layers}

The key innovation of the transformer architecture \cite{vaswani2017attention}, the basis of modern language models, 
is that of (scaled dot-product) (masked) \emph{self-attention}.
This notion of attention can be viewed as particularly useful prescription for how to set attention in a special case of our framework.

Suppose that we are looking at a sequence of $n$ tokens, whose current (pre-transformation) representations at some layer of the model are the vectors $x_1, \ldots, x_n \in \mathbb R^d$. 
The transformation is parameterized by three matrices $W_K, W_Q, W_V \in \mathbb R^{d \times d}$, 
that generate key $k_i = W_K x_i$, query $q_i = W_Q x_i$, and value $v_i = W_V x_i$ representation vectors of each $x_i$.
The output of the transformation are the vectors  $x'_1, \ldots, x'_n$
given by $x_j' = \sum_{i=1}^n \alpha_{i|j} v_i$, where $\alpha_{i|j} := \mathrm{softmax}_i( \langle k_i, q_j \rangle)$.

\newsavebox{\transformerpdg}
\sbox{\transformerpdg}{%
    \def\vdotsmall{\rotatebox{90}{$\cdot{\cdot}\cdot$}}
       \begin{tikzpicture}[center base]
          \node[dpad1] (X1) at (0, 3)  {$X_1$};
          \node             at (0, 2.4) {\vdotsmall};
          \node[dpad1,draw=black] (Xi) at (0, 1.8)  {$X_i$};
          \node             at (0, 0.9)  {\vdotsmall};
          \node[dpad1] (Xn) at (0, 0)  {$X_n$};
          \node[dpad1] (X1p) at (1.8, 3)  {$X'_1$};
          \node              at (1.8, 2.1)  {\vdotsmall};
          \node[dpad1,draw=black] (Xjp) at (1.8, 1.2)  {$X'_j$};
          \node              at (1.8, 0.6) {\vdotsmall};
          \node[dpad1] (Xnp) at (1.8, 0)  {$X'_n$};
          
          \begin{scope}[{arr2,thin,gray}]
          \draw[] (X1) to (X1p);
          \draw[] (X1) to (Xjp); 
          \draw[] (X1) to (Xnp);
          \draw[] (Xi) to (X1p);
          \draw[] (Xi) to (Xnp);
          \draw[] (Xn) to (X1p);
          \draw[] (Xn) to (Xjp);
          \draw[] (Xn) to (Xnp);
          \end{scope}
          \draw[arr2] (Xi) to node[inner sep=2pt,fill=white]{$\p_{ij}$} (Xjp); 
          
          \node at (-0.6,3.4) {$x$};
          \node at (2.4,3.4) {$x'$};
          \begin{scope}[arr2, ->>, shorten <=0pt]
             \draw[] (-0.8,3.5) -- (-0.8, 0) -- (Xn);
             \draw[] (-0.8,3) -- (X1);
             \draw[] (-0.8,1.8) -- (Xi);
    
             \draw[] (2.6,3.5) -- (2.6, 0) -- (Xnp);
             \draw[] (2.6,3) -- (X1p);
             \draw[] (2.6,1.2) -- (Xjp);
          \end{scope}
        \end{tikzpicture}
        }%
\setlength{\columnsep}{1em}
\begin{wrapfigure}{r}{0.44\linewidth}
\begin{center}
    \vspace{-3ex}
    $\dg M(W_{\!K},W_{\!Q},W_V,x,x') $\\
    $:=$\\
    \scalebox{0.95}{\usebox\transformerpdg}
\vspace{-4ex}
\end{center}
\end{wrapfigure}
We can capture this with a relatively simple PDG (right). Start by assembling $2n$ $\mathbb R^d$-valued variables $\mathcal X = \{ X_i \}_{i=1}^n \cup \{ X'_i\}_{i=1}^n$. 
Start with a bipartite graph: for each $(i,j) \in [n]^2$, 
add an arc 
with isotropic Gaussian cpd $\p_{ij} = \mathcal N(X'_j \mid v_i, I)$; finally, 
   add two additional arcs specifying the joint values of $x$ and $x'$.
The resulting PDG
encodes the transformer as the optimal resolution of these many conflicting beliefs. 
 
\begin{linked}{prop}{transformer}
   LIR($\dg M$) in which attention is set to $\varphi_{ij} = \exp \langle k_i, q_j \rangle$, when controlling $x'$, leads to inference in the transformer layer, \ie, 
   $x_j' = \sum_i
      \alpha_{i|j}
     v_i$.
\end{linked}

Notably, this straightforward model shows us that the softmax normalization must be row-wise, which is a common point of confusion; it tells us why attention $\alpha_{i|j}$ is a conditional probability pointing backwards in the network. 
Controlling the parameters $W_K,W_Q,W_V$ trains the layer in a sense, but we do not know whether or not this is equivalent to the standard way of training.

\subsection{Generative Flow Networks}
\label{sec:gfn}

Generative Flow Networks, or GFlowNets \cite{bengio2021flow} aim to draw samples $x \in X$ proportional to a density $R(x)$ by modeling the flow of particles on a directed acyclic graph $(\mathcal S \supseteq X, \mathbb A)$ from a source node $s_0$ to a sink node $s_{\!f}$
    \unskip. 
There are several different ways of parameterizing and training a GFlowNet. 
The most popular approach, called Trajectory Balance (TB) \cite{malkin2022trajectory}, involves parameterizing 
(1) a forward policy $P_F(s' \mid s)$ supported on $(s,s') \in \mathbb A$ that locally chooses edges given current state, 
(2) a policy $P_B(s \mid s')$ that travels backwards along edges, and 
(3) a normalizing constant $Z \approx \sum_{x \in X} R(x)$.
The the training procedure aims to ensure probabilistic consistency between $P_F$ and $P_B$: for every trajectory $\tau = (s_0, s_1, s_n =x, s_{n+1} = s_{\!f})$, we want to ensure that
\vspace{-0.5ex}
\[
   P_F(\tau) := \prod_{i=0}^n P_F(s_{i+1}) 
      \approx \prod_{i=1}^n P_B(s_{i} \mid s_{i+1}) =: P_B(\tau),
\]
where $P_B(\,\cdot \mid s_{\!f}) = R(x)/Z$.
For any distribution $Q(\tau)$ with full support over possible trajectories, 
the quantity
\[
   \mathcal{L}_\mathrm{TB}(Q) := \Ex_{\tau \sim Q}\left[ \log^2 
         \frac{P_F(\tau) Z}{R(x) P_B(\tau\mid x)}  
      \right]
\]
is positive and equals zero iff all constraints are satisfied.
In the special 
case where $Q(\tau) = P_F^\theta(\tau)$ and $P_B$ is fixed, the full gradient
   $\nabla_{\!\theta} \mathcal L_{\mathrm{TB}} =
   \nabla_{\!\theta} \kldiv{P_F^\theta(\tau)}{P_B(\tau)}$ is the same as that of reverse KL \cite{malkin2022gflownets}.
Yet $Q$ need not equal $P_F$ (known as off-policy training), so GFlowNet objectives are thought to extend beyond KL.
Yet by using a certain adaptive refocus strategy, TB can be captured with LIR, which is fundamentally based on KL. 
Concretely, we define a PDG $\dg M(\theta) := $
\newsavebox\gfnpdg
    \sbox\gfnpdg{\begin{tikzpicture}[center base]
       \node[dpad1] (S) at (2.3, 0.6) {$S$};
       \node[dpad1] (Sp) at (2.3, -0.6) {$S'$};
    
       \draw[arr2] (S) to[bend left] node[right]{$P_F^\theta$} (Sp);
       \draw[arr2] (Sp) to[bend left] node[left]{$P_B$} (S);
    
       \node[dpad1] (tau) at (-0.5, 0.5) {$\tau$};
       \node[dpad1] (i) at (0.5, -0.5) {$i$};
       \coordinate (center) at (0.8, 0.3);
    
       \draw[arr2, <-] (tau) to node[above]{$Q!$} +(-1,0);
    
       \draw[arr2] (tau) to node[below left]{$\mathrm{Unif}\{0,1, \ldots, n \}$!} (i);
       \draw[arr2,->>] (tau) to[out=0,in=180] (center) to[out=0,in=-180] (S);
       \draw[arr2,->>] (i) to[out=90, in=180] (center) to[out=0,in=170,looseness=1.8] (Sp);
    \end{tikzpicture}}
\vspace{-1.8ex}
\begin{equation}
    \usebox\gfnpdg
    \label{eqn:tb-gfn}
\end{equation}
where $S$ and $S'$ take possible values $\V S = \V S' = \mathcal S$, $\tau$ ranges over trajectories according to $Q$,  $i$ is an index in the trajectory chosen uniformly at random, and the complicated hyperarc simply ensures that $\tau[i] = S$ and $\tau[i+1] = S'$. 
These are essentially the minimal ingredients needed to put $P_F$, $P_B$, and $Q$ together. 

Now for the key ingredient.
Surprise draws attention; thus, we set the attention in $P_F$ is the forward policy's excess surprisal relative to the baseline $P_B$, that is,
$\varphi(P_F) = \mathrm I_{P_F}[\tau] - \mathrm I_{P_B}[\tau] = \log\frac{P_B(\tau)}{P_F(\tau)}$, 
and vice versa: 
$\varphi(P_B) = \mathrm I_{P_B}[\tau] - \mathrm I_{P_F}[\tau] = \log\frac{P_F(\tau)}{P_B(\tau)}$. 
We then get:

\begin{linked}{prop}{GFN}
\label{prop:gfn}
   The inconsistency of the
   PDG
   $\dg M$ from \eqref{eqn:tb-gfn} with centered surprisal-based attention 
   $\varphi$
   is
   \[
        \aar{\varphi \odot \dg M} 
        =
        \Ex_{\tau \sim Q}\left[
         \frac{1}{|\tau|}
        \log^2 
         \frac{P_F(\tau) Z}{R(x) P_B(\tau\mid x)} 
        \right]
        =: \mathcal L_{\mathrm{ModTB}}.
   \]
   Furthermore, the gradients of $\mathcal L_{\rm ModTB}$ differ only by a factor of two if $\varphi$ is regarded as fixed and not a function of $\theta$. 
   Therefore, LIR on $\dg M$ with attention $\varphi$ and control of $P_F$ (and optionally $P_B$) amounts to training a GFlowNet with ``re-weighted'' TB loss.
\end{linked}

This modified variant of trajectory-balance ($\mathcal L_{\rm ModTB}$), 
is closely related to the original ($\mathcal L_{\rm TB}$); in particular, it has the same global minimum in which all flow constraints are satisfied \citep{bengio2023gflownet}.
That said, there is a meaningful difference in training dynamics induced by the two losses,  because of learning rates and batching.
This is the first instance we know of in which the PDG framework has departed from standard practice for choosing a loss function in an established community. So how does it compare?
As we will see in
\cref{sec:impl-expts} and Appendix~\ref{apx:gfn}, initial experiments suggest that $\mathcal L_{\rm ModTB}$ actually works better than $\mathcal L_{\rm TB}$. For an overview of experimental results, see \cref{fig:gfn}.

\subsection{Decision Making}
In this final illustration of the 
expressive power of LIR, we return to psychological models of agents,
as we show how some standard decision rules can be viewed as inconsistency minimization. 
PDG inconsistencies can represent expected costs \cite[\S 4, \S6]{oli-dissertation},
and thus can jointly represent an agent's probabilistic beliefs and utilities.
What does it mean to make decisions to as to minimize inconsistency in this context? 

Let's formalize the traditional setup of decision theory with a PDG. 
Suppose we are trying to choose an action (a setting of the variable $A$),
$S$ but we are uncertain about the current state of the world (captured by a variable $S$). Let $O$ be a variable representing the final outcome.
To connect these pieces of the puzzle, the standard assumption is that 
we have 
\begin{enumerate}[nosep]
\item 
a belief $p(S)$ about the state of the world,
\item  a utility function $u : \V O \to \mathbb R$ on outcomes,
and 
\item 
an understanding $\tau(O \mid S,A)$ of how our action $A$ and the state $S$ determine the outcome $O$. 
\end{enumerate}
It is easy to encode this information in a PDG:
\vspace{-1ex}
\begin{equation}
   \begin{tikzpicture}[scale=0.8,center base]
    \begin{scope}
       \node[dpadded] (S) at (-0.4,0) {$S$};
       \draw[arr2, <-] (S) -- node[above]{$p$} ++(-1.5, 0); 

       \node[dpadded] (O) at (2,0) {$O$};
       \node[dpadded] (U) at (4,0) {$U$};
       \draw[arr1, ->>] (O) -- node[above, pos=0.35]{$u$}  (U);
       \node[dpadded] (A) at (0.5,1.3) {$A$};
           \mergearr[arr1,->] SAO
       \node[below=1pt of center-SAO]{$\tau$};
   \end{scope}
   \end{tikzpicture}~\raisebox{-1em}{.}
   \label{pdg:action}
\end{equation}
\vspace{-1.5ex}

What's missing is the idea 
that a higher numerical utility should be ``better'' than a lower one. 
For this, we can add a ``soft constraint'' \citep[see][\S4.2.2]{oli-dissertation}.
\vfull{\unskip\footnote{%
    at a technical level, this can be done by 
    (1) including a variable $\Tru$ with possible values $\V \Tru = \{ \truf, \trut\}$ but is constrained to $\Tru=\trut$,
    and  
    (2) adding a cpd $b(\Tru \mid U)$ that threatens to violate that constraint $\Tru{=}\trut$ with probability decreasing in $U$. 
}}
Let 
$\dg M$
represent the PDG above augmented with such a soft constraint. 

To give a stylized interpretation, imagine that there is a part of you that you cannot control called ``Faith''.
Faith is optimistic: 
   she disbelieves outcomes that are undesirable,
   creating epistemic conflict in proportion to disutility.
If you have no control over Faith, but still have to pay attention to her,
then selecting actions to minimize inconsistency
is a decision rule that interpolates between
\begin{enumerate}[nosep,label={(\alph*)}]
   \item 
   maximizing expected utility (when $\beta_p \gg \beta_b$), and
   \item 
   maximizing maximum utility (when $\beta_p \ll \beta_b$),
\end{enumerate}
where  $\beta_p$ is your confidence in your prior probabilistic belief $p(S)$, and $\beta_b$ is the attention to the soft constraint $b$ (\ie, your ``degree of faith'').

\begin{linked}{prop}{eumaxmax}
   Let $\dg M$ be the PDG in \eqref{pdg:action}
   together with the cpd $b(\Tru{=}\trut \mid U{=}u) = k \cdot \exp( u )$ for some $k > 0$. 
   \begin{enumerate}[nosep,label={(\alph*)}]
       \item 
   If $\beta_p = \infty$ and $\beta_b < \infty$, then 
   the action(s) $a \in \V\! A$ that minimize the inconsistency
   are those that maximize expected utility.
   Formally,
   for all $\gamma \ge 0$, 
   \[
       \argmin_{a \in \V\!A} \aar[\big]{\dg M
            + A{=}a}_\gamma 
       = \argmax_{a \in \V\! A} 
           \smash{\Ex_{\substack{s\sim p \\ o \sim \tau|s,a}}} \big[ u(o) \big].
   \]
   \item 
   If $\beta_p < \infty$ and $\beta_b = \infty$, then 
   the action(s) $ a \in \V\! A$ that minimize overall inconsistency are
   those that can lead to the best possible outcome, \ie, 
   \[
       \argmin_{a \in \V\!A} \aar[\big]{\dg M
            + A{=}a}_\gamma 
       = \argmax_{a \in \V\! A} \;
           \max_{s \in \V S} \;
           \Ex_{o \sim \tau|s,a} [ u(o) ].
   \]
   \end{enumerate}
\end{linked}

We suspect that it may also be possible to implement other decision rules such as minimax or maximin.
Like the GAN objective however, these decision rules look like two-player games, and so will likely require two different focuses of LIR, rather than just one. 
We leave a careful exploration of this avenue to future work.

We conclude with a high-level observation. 
One's preferences, beliefs, and actions can be jointly modeled with a PDG; when 
when that PDG is inconsistent,
there are three possible resolutions (just as in \cref{sec:classification}).
You can 
resolve the inconsistency by changing your action, 
   which amounts to maximizing expected utility (\cref{prop:eumaxmax}.1);
   this is thought of as the {rational} approach.
Alternatively, 
   you can change your preferences so as to become content with your current situation, 
   which is perhaps a more zen perspective. 
These two resolutions of the internal conflict are two halves of the famous Serenity Prayer \citep{serenity-prayer}:
\begin{quotation}\it\small
   \noindent
   Give us the courage to change what must be altered, serenity to accept what can not be helped, and insight to know the one from the other.
\end{quotation}
\vspace{-1ex}
That ``insight'' amounts to the choice of the control mask $\chi$. 
There is also a third way to resolve the inconsistency: change beliefs $p(S)$ about the state of the world (\ie, wishful thinking).
However, such a strategy may lead to more inconsistency in the future, making it undesirable and ineffective.  
This example clearly shows that
   there can be good reasons for 
   restricting control $\chi$ 
   beyond computational savings or the limits of one's power.

\vspace{-0.5ex}

\section{Implementation \& Experiments}
    \label{sec:impl-expts}

\begin{figure*}[t!]
    \centering
    \includegraphics[width=0.8\textwidth]{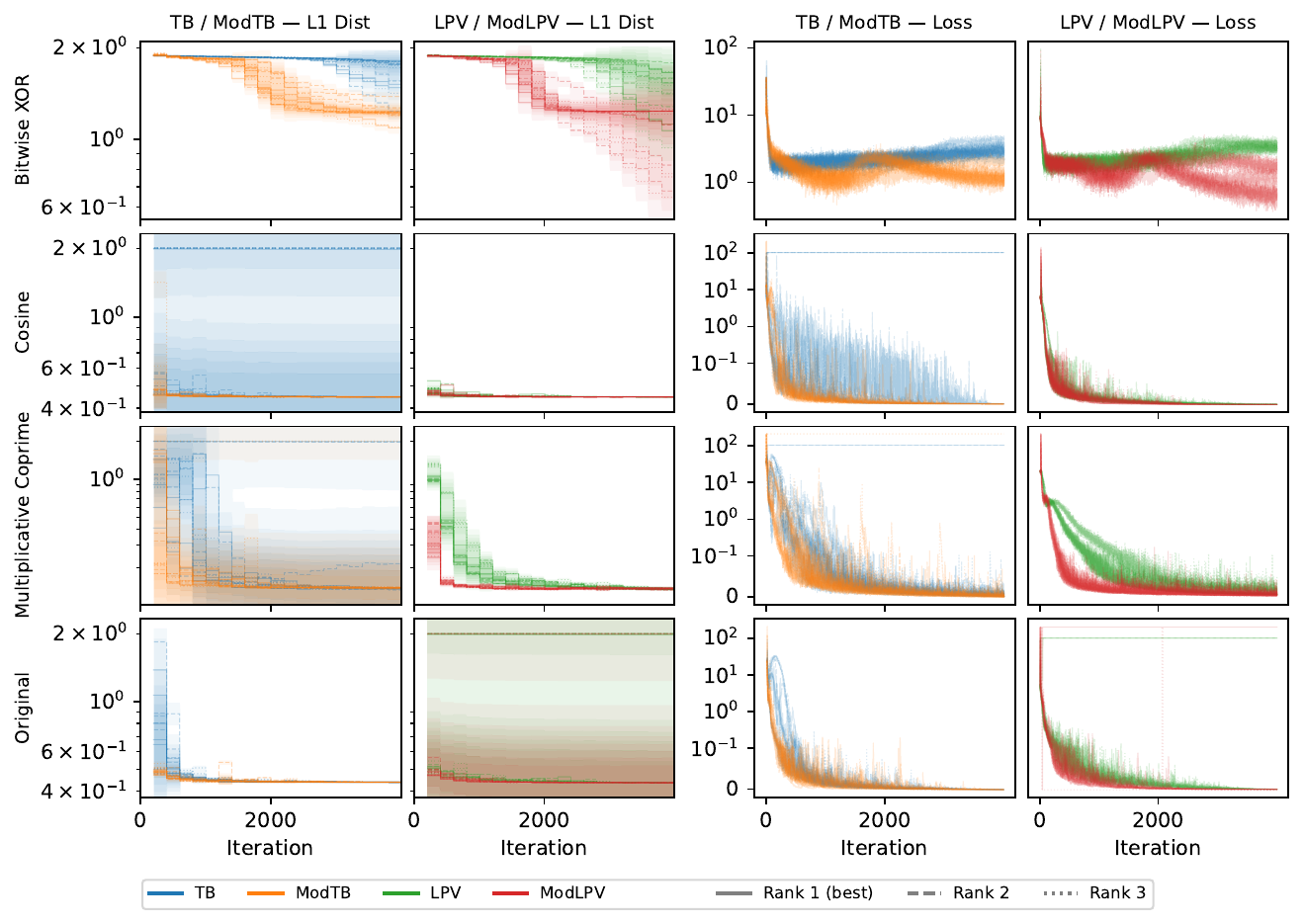}
    \caption{\small Performance (L1 distance between the estimated and true posterior distribution; left) and loss (computed using the unnormalized variant at evaluation time for visualization at the same scale; right) of the Trajectory Balance (TB) and Log-Partition Variance (LPV) losses and their normalized variants (Mod TB and Mod LPV, respectively). Figures show traces from the 3 top hyperparameter configurations (ranks) found with 5 seeds each rank. The performance curves show overlaid on the traces a 1D KDE fit in log space at each timestep, showing relative per-step density.}
    \label{fig:gfn}
    \vspace{-3.5mm}
\end{figure*}

\textbf{Implementation.}
We implemented LIR by extending an existing code-base for PDGs. The primary challenge was in getting (sufficiently high-quality) gradients of the inconsistency efficiently enough to run the procedure on small examples and use an ODE solver. 
More details can be found in Appendix~\ref{sec:impl-details}.

\textbf{Synthetic Experiments.}
As a proof-of-concept of our generic LIR implementation, we evaluate it by comparing the effect of various \textsc{Refocus} procedures on a dataset of randomly generated PDGs.
We compare the global uniform attention, in which  all arcs are active ($\bbeta = \mat 1$), to various more local foci, in which only a subset of edges are active ($\beta(a) = 1$ for some arcs $a$ and $\beta(a') = 0$ for others).
In all, we study four refocus strategies: global attention (all arcs active), local (half of arcs randomly deactivated), node-neighborhood (arcs active if they are sources or targets of a a randomly chosen node), and random smooth masks ($\bbeta$ drawn from an exponential distribution).
In each case, we evaluate the behavior of LIR based on several criteria:
(1) the (rate of) reduction of inconsistency from initialization;
(2) the computational cost;
(3) the degree to which the optimal distribution $\mu^* \in \arg\min \OInc_{\dg M}(\mu)$ is distorted during the process. 
Details about these experiments, can be found in Appendix~\ref{sec:expt-details}.

\paragraph{Modified TB Losses.} Arising from our observations in \ref{sec:gfn}, we noted that the original TB loss omits a trajectory length scaling factor (see \cref{prop:gfn} for the definition of $\mathcal L_{\rm ModTB}$, the Modified TB loss). We analyze whether including this scaling factor during training produces meaningfully different results under four HyperGrid environments of varying difficulty. See Appendix \ref{apx:gfn} for all experimental details.

A cousin of TB is the Log Partition Variance loss, or "VarGrad" \cite{richter2020vargrad,malkin2023gflownets}, which admits an analogous modified variant. 
For a minibatch of trajectories $(\tau_i)_{i=1}^m$ drawn from $Q$, let 
$$s_i := \log \frac{P_F(\tau_i)}{R(x_i)\, P_B(\tau_i \mid x_i)}, \quad \bar{s} = \frac{1}{m}\sum_{i} s_i$$ be the mini-batch mean, and
and define
\begin{align*}
\mathcal{L}_{\text{ModLPV}} = \frac{1}{m}\sum_{i=1}^{m} \frac{(s_i - \bar{s})^2}{|\tau_i|}.
\end{align*}
Both were implemented using \texttt{torchgfn} \cite{viviano2026torchgfnpytorchgflownetlibrary}. Figure \ref{fig:gfn} shows 5 seeds from the top 3 hyperparameter settings found after a 50 Optuna trials \cite{optuna}. All models covered all modes in  all environments, but training stability varied, with larger differences apparent in the more difficult compositional environments 
The modified losses generally show faster convergence at the earlier stages of training. For the challenging environments (\emph{\eg}, \textit{Bitwise XOR}), there was a substantial difference also in the fidelity of the learned distribution.
Of note, the best gradient clipping (norm-based) value found for the best LPV models was much more aggressive (see Table \ref{tab:gfn}), while the modified variant was much more tolerant of larger gradient norms. See Appendix \ref{apx:gfn} Table \ref{apx:gfnrank1} for more.
\begin{table}[ht]
  \centering
  \small
  \begin{tabular}{lccc}
    Environment & LPV & ModLPV & Ratio \\
    \midrule
    Original               & 0.025 & 0.23 & $\sim10\times$ \\
    Cosine                 & 0.10  & 0.045 & $\sim2\times$ \\
    Bitwise XOR            & 0.033 & 1.19 & $\mathbf{\sim36\times}$ \\
    Multiplicative Coprime & 0.013 & 0.84 & $\mathbf{\sim65\times}$ \\
    \bottomrule
  \end{tabular}
    \caption{\small Median gradient clipping threshold (norm-based) across the top-10 Optuna trials for each loss.}
  \label{tab:gfn}
  \vspace{-3mm}
\end{table}

\vspace{-0.5ex}
\section{Conclusion}

We have introduced Local Inconsistency Resolution (LIR), a framework that unifies and explains a wide variety of important algorithms, as stemming from different allocations of attention and control. 
It has led us to a slightly better formulation of a standard loss, in the world of GFlowNets.

Yet many fundamental questions remain.
Under what assumptions does LIR converge? 
Can we learn adaptive \textsc{Refocus} mechanisms (drawing inspiration from the transformer architecture)?
Can we use it to discover new hybrid algorithms?
Answering these questions could advance a new kind of intelligent system---one that is more interpretable and potentially safer: behavior emerges from resolving conflicts among explicit adjustable beliefs rather than rigidly pursuing a fixed goal in the world.

\subsubsection*{Acknowledgments}
This work was supported in part by
funding from NSERC and CIFAR. 

\bibliography{lir-refs}

\clearpage
\section*{Checklist}

\begin{enumerate}

  \item For all models and algorithms presented, check if you include:
  \begin{enumerate}
    \item A clear description of the mathematical setting, assumptions, algorithm, and/or model. 
    \textbf{Yes.}
    \item An analysis of the properties and complexity (time, space, sample size) of any algorithm. 
    \textbf{No.}
    \item (Optional) Anonymized source code, with specification of all dependencies, including external libraries. 
    \textbf{Yes.}
  \end{enumerate}

  \item For any theoretical claim, check if you include:
  \begin{enumerate}
    \item Statements of the full set of assumptions of all theoretical results. \textbf{Yes.}
    \item Complete proofs of all theoretical results. \textbf{Yes.}
    \item Clear explanations of any assumptions. \textbf{Yes.}
  \end{enumerate}

  \item For all figures and tables that present empirical results, check if you include:
  \begin{enumerate}
    \item The code, data, and instructions needed to reproduce the main experimental results (either in the supplemental material or as a URL). \textbf{At least some.}
    \item All the training details (\eg, data splits, hyperparameters, how they were chosen). \textbf{Yes.}
    \item A clear definition of the specific measure or statistics and error bars (\eg, with respect to the random seed after running experiments multiple times). \textbf{Yes.}
    \item A description of the computing infrastructure used. (\eg, type of GPUs, internal cluster, or cloud provider). \textbf{Yes.}
  \end{enumerate}

  \item If you are using existing assets (\eg, code, data, models) or curating/releasing new assets, check if you include:
  \begin{enumerate}
    \item Citations of the creator If your work uses existing assets. \textbf{Not Applicable}
    \item The license information of the assets, if applicable.\textbf{Not Applicable}
    \item New assets either in the supplemental material or as a URL, if applicable. \textbf{Not Applicable}
    \item Information about consent from data providers/curators. \textbf{Not Applicable}
    \item Discussion of sensible content if applicable, \eg, personally identifiable information or offensive content. \textbf{Not Applicable}
  \end{enumerate}

  \item If you used crowdsourcing or conducted research with human subjects, check if you include:
  \begin{enumerate}
    \item The full text of instructions given to participants and screenshots. \textbf{Not Applicable}
    \item Descriptions of potential participant risks, with links to Institutional Review Board (IRB) approvals if applicable. \textbf{Not Applicable}
    \item The estimated hourly wage paid to participants and the total amount spent on participant compensation. \textbf{Not Applicable}
  \end{enumerate}

\end{enumerate}

\clearpage
\appendix
\onecolumn

\aistatstitle{Local Inconsistency Resolution: The Interplay between Attention and Control in Probabilistic Models \\[1ex]
Supplementary Materials}

\section{Experimental and Implementation Details}

\subsection{Implementation}
    \label{sec:impl-details}

We use PyTorch for our implementation. One of the most basic PDG computations is to compute the incompatibility, denoted by $f$, of a PDG $\dg M$ given joint distribution $\mu$ and attention mask $\varphi$. The attention mask decides how much each arc contributes to the incompatibility. As a result, many variables may have no contribution at all. We prune these variables for computational efficiency.

The $\gamma$-inconsistency of a PDG $\dg M$ (\cref{eq:inconsistency}) is obtained by finding the joint distribution $\mu$ that minimizes incompatibility. We represent the joint distribution $\mu$ as a point on a simplex and use the Adam optimizer for minimizing incompatibility w.r.t. $\mu$. This inner optimization procedure is warm-started from the last computed joint distribution to increase computational efficiency.

To obtain the gradient of inconsistency w.r.t. learnable parameters, denoted by $\theta$, we use the \emph{envelope theorem} to simplify differentiation through the inner optimization problem
\[
\nabla_\theta \min_\mu f(\mu,\theta)
\;=\;
\nabla_\theta f(\mu^*(\theta),\theta),
\]
where $\mu^*(\theta)=\argmin_\mu f(\mu,\theta)$. Hence, there is no need to backpropagate through the inner optimization:
we simply plug in the numerically optimized joint distribution \(\mu^*\) and compute
gradients of the incompatibility w.r.t. $\theta$. This gradient is then scaled by the control mask $\chi$. For the outer optimization of $\theta$ we implement both standard optimizers such as Adam as well as ODE solvers such as Runge-Kutta.

\subsection{Synthetic Experiments}
    \label{sec:expt-details}
Local inconsistency resolution employs attention mechanisms to selectively focus computational resources on subsets of the model during training. However, the choice of refocus strategy fundamentally impacts both efficacy of inconsistency resolution and the computational efficiency of the learning process. To systematically evaluate different attention strategies, we designed a suit of synthetic experiments using PDGs with controllable structural properties and known inconsistencies. 

We generate synthetic PDGs with deliberately introduced inconsistencies to create a controlled experimental environment. Each PDG is constrained as follows:
\paragraph{Variables} Each PDG contains $n$ discrete random variables $\{X_1, \ldots, X_n\}$, where each variable $X_i$ has a small finite domain $\V\!X_i$ (consisting of 2 or three possible values).

\paragraph{Graph Structure}
For these preliminary experiments, we study a simple class of discrete PDGs whose underlying hypergraph structure is an ordinary directed graph.
The edge set is constructed in two phases: 
\begin{enumerate}
    \item \textbf{Base Chain Construction} We first create a chain structure with $\lfloor m/2 \rfloor$ edges, where $m$ is the target number of edges:
    \begin{equation*}
        \Ar_{\text{chain}} = \{(i \to i+1) : i \in [1, \lfloor m/2 \rfloor]\}
    \end{equation*}
    \item \textbf{Conflict Edge Addition:} We then add additional edges preferentially targeting nodes that already have incoming edges, creating conflict points:
    \begin{equation*}
        \Ar_{\text{conflict}} = \{(i \to j) : j \in \text{Targets}(\Ar_{\text{chain}}), i \neq j\}
    \end{equation*}
\end{enumerate}
This construction guarantees that certain nodes receive multiple incoming edges with potentially conflicting conditional probability specifications, ensuring non-zero inconsistency.

\paragraph{Conditional Probability Distributions} For each edge $(i\to j) \in \Ar$, we assign a randomly initialized conditional probability table $\p_{ij}(X_j | X_i)$ drawn uniformly from the probability simplex. Crucially, these CPDs are \textit{not} jointly consistent, there generally does not exist a joint distribution $\mu$ that satisfies all CPDs simultaneously, guaranteeing measurable inconsistency.

We evaluate four PDG configurations, denoted as $\texttt{chain\_}(n)\texttt{v\_}(m)\texttt{e}$ for $n$ variables and $m$ edges.

\begin{center}
\begin{tabular}{lccc}
\hline
\textbf{PDG} & \textbf{Variables} & \textbf{Edges} & \textbf{Initial Inconsistency} \\
\hline
$\texttt{chain\_4v\_3e}$ & 4 & 3 & 0.159294 \\
$\texttt{chain\_5v\_4e}$ & 5 & 4 & 0.168060 \\
$\texttt{chain\_6v\_5e}$ & 6 & 5 & 0.209741 \\
$\texttt{chain\_7v\_6e}$ & 7 & 6 & 0.145324 \\
\hline
\end{tabular}
\end{center}

\subsubsection{Inconsistency Measure}

As is the case for most of our results, our 
Given a PDG $\dg M$ with edge set $\Ar$ and a joint distribution $\mu$ over variables $\X = (X_1, \ldots, X_n)$, recall (from \eqref{eq:oinc}) that the (observational) incompatibility is measured as:
\begin{equation}
    \OInc(\mu, \dg M) = \sum_{(i,j) \in \Ar} \beta_{ij} \cdot D_{\text{KL}}\left(\mu(X_j | X_i) \,\|\, \p_{ij}(X_j | X_i)\right),
\end{equation}%
where: %

\begin{itemize}[nosep,itemsep=0.1ex]
    \item $\mu(X_j | X_i)$ is the conditional distribution induced by the joint distribution $\mu$
    \item $\p_{ij}(X_j | X_i)$ is the conditional probability table specified by edge $i \to j$
    \item $\beta_{ij} \in [0,1]$ is the attention weight for edge $i \to j$
    \item $D_{\text{KL}}(P \| Q) = \kldiv PQ = \sum_x P(x) \log \frac{P(x)}{Q(x)}$ is the Kullback-Leibler divergence.
\end{itemize}
The optimal joint distribution $\mu^*$ for a given set of CPDs is obtained by solving:
\begin{equation}
    \mu^* = \argmin_{\mu \in \Delta \V\!\X} \OInc(\mu, \dg M),
\end{equation}
and the \emph{inconsistency} of $\dg M$ is given by
\begin{equation}
    \aar[\big]{\dg M} = \OInc(\mu^*, \dg M) = \inf_{\mu \in \Delta\V\!\X} \OInc(\mu, \dg M).
\end{equation}

\subsubsection{Refocus Strategies}

We evaluate three refocus strategies that control the $\beta$ weights.
\vspace{-0.4cm}
\paragraph{Uniform Focus}
All edges receive full attention at all times:
\begin{equation}
    \beta_{ij}^{\text{uniform}} = 1 \quad \forall (i \to j) \in \Ar
\end{equation}
This strategy optimizes all edges simultaneously, maximizing global consistency but requiring the most computational resources.
\vspace{-0.4cm}

\paragraph{Partial Focus}
At each training iteration $t$, we randomly select a subset $\Ar_{\text{active}}(t) \subseteq \Ar$ with $|\Ar_{\text{active}}(t)| = \lfloor |\Ar|/2 \rfloor$, and focus attention only on this subset:
\begin{equation}
    \beta_{ij}^{\text{partial}}(t) = \begin{cases}
        1 & \text{if } (i\to j) \in \Ar_{\text{active}}(t) \\
        0 & \text{otherwise}
    \end{cases}
\end{equation}
This strategy reduces computational cost by ignoring half the graph at any given time, but may achieve slower or incomplete convergence.
\vspace{-0.4cm}

\paragraph{Hub Focus}
At each iteration $t$, we select a random focus node $X_{\text{focus}}(t)$, and attend to all edges incident to this node:
\begin{equation}
    \beta_{ij}^{\text{hub}}(t) = \begin{cases}
        1 & \text{if } X_i = X_{\text{focus}}(t) \text{ or } X_j = X_{\text{focus}}(t) \\
        0 & \text{otherwise}
    \end{cases}
\end{equation}
This strategy focuses on potential inconsistencies that are, in a sense,  localized to a single variable. 

\subsubsection{Training Procedure}
For each PDG and focus strategy, we perform the following training procedure:
\begin{enumerate}
    \item \textbf{Initialization:} Convert each fixed CPD to a learnable parameterized conditional probability distribution (\verb|ParamCPD|) initialized from the original CPD values.
    
    \item \textbf{Initial Joint Distribution:} Compute the initial optimal joint distribution $\mu_{\text{init}}^*$ by solving:
    \begin{equation}
        \mu_{\text{init}}^* = \argmin_{\mu} \OInc(\mu, \dg M_{\text{init}})
    \end{equation}
    using the Adam optimizer with $\gamma = 0$ (no entropy regularization) for 50 iterations.
\item \textbf{LIR Training:} Apply LIR with the specified refocus strategy for $T = 20$ timesteps.
At each timestep $t$, our implementation of LIR updates CPD parameters $\theta$ by approximating the solution to the ODE
\[
    \theta_{t+1} \gets \mathrm{SolveODE}\Big[ \dot\theta =  \nabla_\theta \aar[\big]{\dg M(\theta), \bbeta}; \;\verb|init|=\theta_t \Big]\]
by applying \verb|#outer_iterations| gradient-based steps of learning rate $\eta$. (We have effectively set a uniform control mask $\chi=\eta$ equal to the learning rate.)
The attention weights $\bbeta$ are determined by the chosen focus strategy. 
In this particular experiment, that outer gradient step is done by the Adam optimizer. 

Within each outer iteration $o = 1, 2, \cdots$, we approximate the optimal joint distribution $\mu^*_{(t,o)}$ by fine-tuning the previous solution $\mu^*_{(t,o)}$ for \verb|#inner_iterations| steps. 
For more details, see \cref{sec:impl-details} and the code in \verb|lir_simpler.py|.

Our experiments use the following values:
\begin{itemize}
    \item 10 outer iterations per timestep to approximate the ODE solution
    \item 20 inner iterations for optimizing $\mu^*$ per outer iteration
    \item Learning rate $\eta = 0.05$
\end{itemize}

    \item \textbf{Final Joint Distribution:} After training, compute the final optimal joint distribution $\mu_{\text{final}}^*$ using the updated CPDs, for the purposes of analysis. 
\end{enumerate}

\subsubsection{Evaluation Metrics}

We evaluate each refocus strategy using three complementary metrics:

\paragraph{Resolution Percentage}

The resolution percentage measures the reduction in inconsistency:
\begin{align}
    \text{Resolution} &= \frac{\OInc(\mu_{\text{init}}^*, \dg M_{\text{init}}) - \OInc(\mu_{\text{final}}^*, \dg M_{\text{final}})}{\OInc(\mu_{\text{init}}^*, \dg M_{\text{init}})} \times 100\% 
    &= \frac{\aar{ \dg M_{\text{init}}} - \aar{ \dg M_{\text{final}}}}{\aar{ \dg M_{\text{init}}} } \times 100\%
\end{align}

Higher resolution indicates more effective inconsistency reduction.
Fig~\ref{fig:compare-initial-final} compares the initial and final inconsistency values across multiple PDGs and three inconsistency-resolution refocus strategies: global focus, partial focus, and hub focus. Across all PDGs, inconsistency values consistently decreased after training, confirming that the optimization procedure effectively resolved contradictions in the dependency structure. The magnitude of reduction varied by both graph complexity and refocus procedure: 
\begin{figure}[t]
    \centering
    \includegraphics[width=\linewidth]{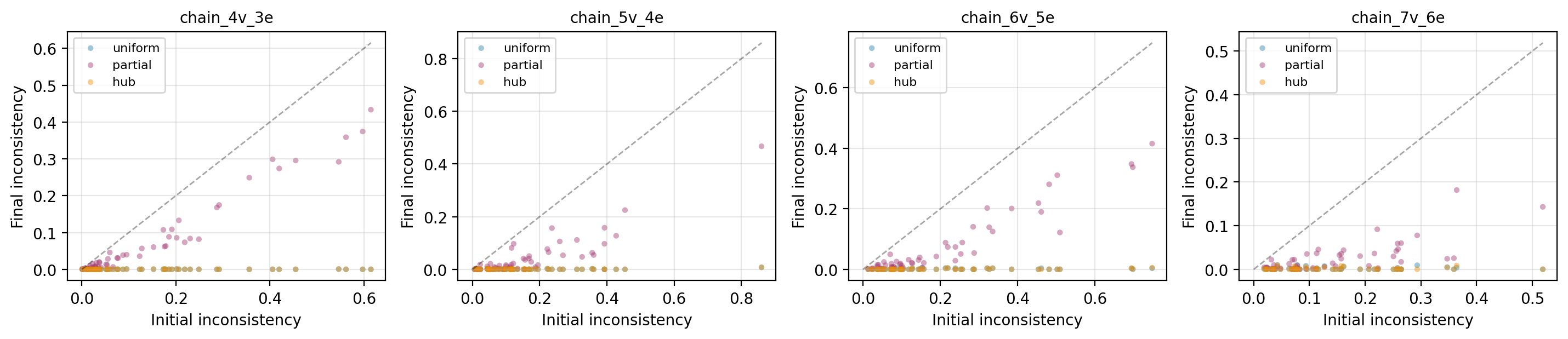}
    \caption{Initial vs. final inconsistency across PDGs for \textit{uniform}, \textit{partial}, and \textit{hub} focus strategies.}
    \label{fig:compare-initial-final}
\end{figure}

\paragraph
{Distortion Analysis: Measuring Changes to the Joint Distribution}
There is a simple but inadvisable and destructive way to instantly resolve all inconsistencies: set all beliefs to uniform (or delete them altogether). Does something like this happen in our case?
While resolution percentage quantifies how effectively a refocus strategy reduces inconsistency, it does not capture how much the underlying probabilistic beliefs change during the resolution process. Two strategies might achieve similar resolution but differ substantially in the magnitude of changes they induce to the optimal joint distribution $\mu^*$. To address this, we introduce a {distortion metric} that measures the distance between the initial and final joint distributions.
We employ the \emph{Total Variation (TV) distance}, a standard metric for comparing probability distributions, defined as:

\begin{equation}
\label{eq:tv_distance}
D_{TV}(\mu_{\text{init}}^*, \mu_{\text{final}}^*) = \frac{1}{2}\sum_{\mathbf{x} \in \V\!\X} \left|\mu_{\text{init}}^*(\mathbf{x}) - \mu_{\text{final}}^*(\mathbf{x})\right|
 = \max_{A \subseteq \V\!\X} \big| \mu_{\text{init}}^*(A) - \mu_{\text{final}}^*(A) \big|.
\end{equation}

\begin{figure}
    \centering
    \includegraphics[width=\linewidth]{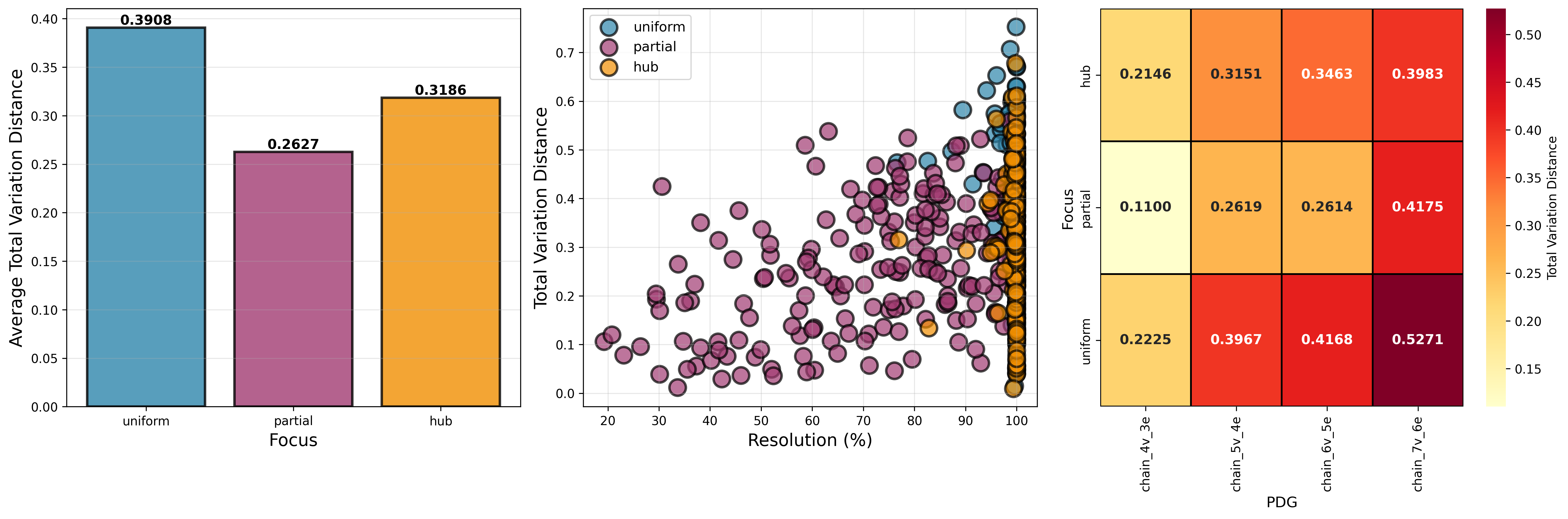}
    \caption{A breakdown of TV distortion through the LIR process across refocus strategies and PDGs}
    \label{fig:distortion}
\end{figure}

The reported TV distortion in \cref{fig:distortion} is the distance between initial $\mu_{\text{init}}$ and final optimized $ \mu_{\text{final}}$, not a regularized penalty for cpd change. So strategies that resolve inconsistency more aggressively can naturally produce larger mu shifts, and therefore larger TV distance. 

As seen in \cref{fig:distortion}, the hub refocus strategy induces higher distortion than alternatives, particularly for the larger PDGs. The partial focus strategy appears to induce less distortion, but in fact produces more distortion per unit of inconsistency resolution.
All results presented here have additional distortion owing to the optimizer (here we've used Adam instead of a dedicated ODE solver), and the fact that we do not pre-multiply by the inverse Fisher matrix to get natural gradients. 

\paragraph{Resolution Performance}

\begin{table}[h]
\centering
\caption{Inconsistency resolution performance by refocus strategy}
\label{tab:resolution}
\begin{tabular}{lcccc}
\hline
\textbf{Refocus Strategy} & \textbf{Avg Resolution (\%)} & \textbf{Avg TV Dist.} \\
\hline
Uniform        & 99.23 & 0.3908 \\
Partial         & 72.19 & 0.2627 \\
Hub    & 99.37 & 0.3186 \\
\hline
\end{tabular}
\end{table}

\begin{figure}[t]
    \centering
    \includegraphics[width=\linewidth]{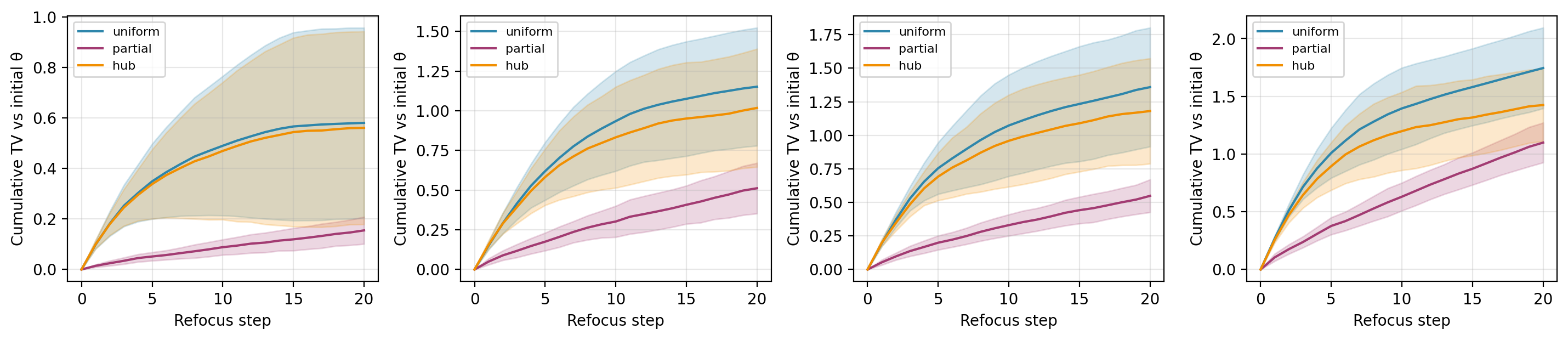}
    \caption{Mean cumulative total variation distance between current CPD parameters ($\theta$) and their initial values ($\theta_0$) across LIR refocus steps, shown separately for each PDG and refocus strategy (uniform, partial, hub), with shaded bands indicating ±1 standard deviation over CPD replicates.}
    \label{fig:heatmap}
\end{figure}

\cref{fig:heatmap} presents a resolution percentage heatmap comparing the effectiveness of three inconsistency-resolution focus strategies (uniform, partial, and hub) across four PDGs. The color scale represents the percentage reduction in inconsistency, with higher values (green) indicating faster convergence. Across all PDGs, both uniform and hub focus consistently achieved high resolution percentages, particularly in larger graphs such as \texttt{chain\_7v\_6e}, where both surpassed $90\%$. Partial focus, however, displayed considerably weaker performance in smaller PDGs. 

\begin{figure}[htbp]
    \centering
    \includegraphics[width=\linewidth]{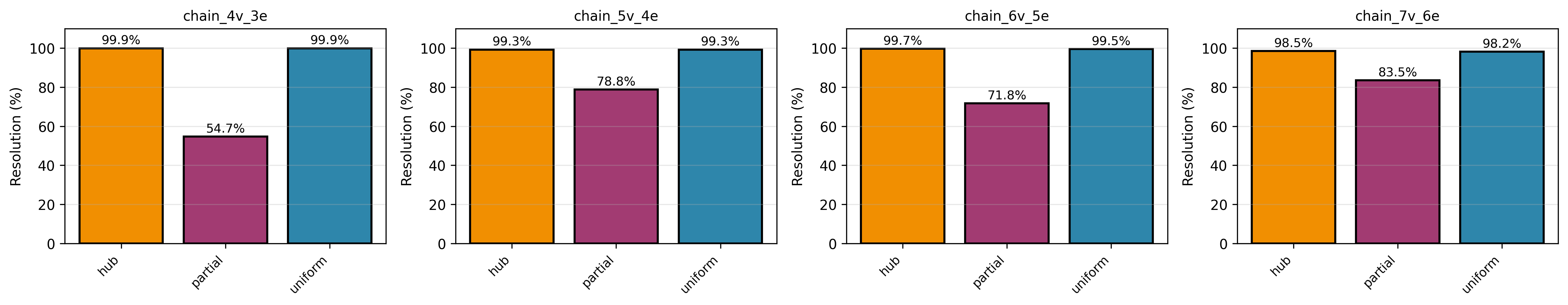}
    \caption{Resolution performance by focus strategy across four PDGs. Uniform and hub approaches consistently achieve higher inconsistency reduction than local updates, with performance converging at larger graph sizes.}
    \label{fig:four_panels}
\end{figure}

Fig~\ref{fig:four_panels} summarizes the resolution performance of the three inconsistency-resolution strategies—uniform, partial, and hub across four PDGs. Across all PDGs, the uniform and hub strategies consistently outperform the partial approach.

The partial focus strategy catches up somewhat in larger graphs (up to $83.5\%$), suggesting that localized updates may become more effective when many more interdependencies exist. Overall, these results indicate that the uniform and hub focus may scale better across larger PDGs (at least with this kind of topology), while partial focus is typically less effective, and sensitive to graph density and connectivity.

\section{Details on Belief Propagation} \label{sec:bp-details}

We now define the foci.
Modulo a small subtlety,
the following is essentially true:
\eqref{eq:X->a}
selects
$C_{X \sto a} := \{ m_{X \sto a} \}$ so as to
minimize 1-inconsistency in  context
$A_{X \sto a} := \{ m_{b \sto X} \}_{b \in \partial X \setminus a} \cup \{m_{X \sto a}\}$,
while
\eqref{eq:a->X}
selects
$C_{a \sto X} := \{ m_{a \sto X} \}$
so as to minimize the 1-inconsistency in
context
$A_{a \sto X} := \{ \phi_a, m_{a \sto X} \} \cup \{ m_{Y \sto a} \}_{Y \in \mat X_a \setminus X}$.

The only wrinkle is that we do not attend to
   the \emph{structural} aspect of the edges $e$ that we're updating;
       \ie, we must select $\varphi$ to effectively set $\alpha_e = 0$.
Intuitively: although all of the input messages summarize causal information,
   we're trying to capture that information with a distribution.
   Thus, it's not appropriate to attend to the causal structure of the edges that we're modifying. 
Thus, for each $f \in\bigcup_{a\in \Ar,X \in \mat X_a}\{a \sto X, X \sto a, X\}$, define
a focus $(\varphi_f, \chi_f) \in \mat F$ according to
\[
   \varphi_f(a) :=  \begin{cases}
       \binom 11 & \text{ if } a \in A_{f} \setminus C_{f} \\
       \binom 10 & \text{ if } a \in C_{f} \\
       \binom 00 & \text{otherwise}
   \end{cases},
   \qquad\qquad
   \chi_f(a) := \begin{cases}
   \infty & \text{ if } a \in C_{v} \\
   0 & \text{ otherwise}.
\end{cases}
\]
where $\binom{\phi_1}{\phi_2}$ scales $\beta_a$ by $\phi_1$ and $\alpha_a$ by $\phi_2$.
With these definitions, \cref{prop:bp} follows easily.

\section{Note on Densities, Mass Functions, and Constants}
    \label{sec:density-note}

Before we move on to the proofs, we start by explicitly addressing an (ultimately inconsequential) slight-of-hand. 
While $\kldiv{\delta_x(X)}{p(X)} = -\log p(X{=}x)= \infty$ when $p(X{=}x) = 0$, as is the case when $p$ is the measure corresponding to a positive density on real numbers (such as a normal distribution), 
there is an important sense in which it effectively equals $- \log p(x) + \mathrm{const}$, where $p(x) = \frac{\mathrm d p}{\mathrm d\lambda}(x)$ is the density with respect to the base (Lebesgue) measure $\lambda$. 
For example, if we discretize the space $\V\! X$ of values of $X$ as a countable collection of voxels of size $\epsilon$, resulting in a new discrete variable $X_\epsilon$ and a cpd $p^\epsilon$ over its values, then 
\[
    \kldiv{\delta_x(X_\epsilon)}{p^\epsilon(X_\epsilon)}
        = - \log p(x) + o(\epsilon^2) + \log \frac1\epsilon.
\]
We obtain essentially the same limit if we use a definition of the Dirac distribution $\delta_x(X)$ as a limit of normal distributions whose variance goes to zero. 
In our case, since we are only ever interested in the gradient of the inconsistency with respect to parameters $\theta$, and $\nabla_\theta \log \frac1\epsilon = 0$, so $\lim_{\epsilon \to 0} \nabla_\theta \kldiv{\delta_x(X_\epsilon)}{p^\epsilon(X_\epsilon)} = - \nabla_\theta \log p_\theta(x)$.

We therefore perform the remainder of our calculations as though
$\kldiv{\delta_x(X)}{p^\theta(X)}
        = -%
        \log p_\theta(x)$.

The same slight-of-hand is standard practice in variational inference for essentially these reasons.
For example, suppose we have a latent variable model $p(X,Z)$ where $X$ and $Z$ are both real-valued variables that we measure in meters; the density $p(x,z)$ is then probability per square meter. In the expression $\textrm{ELBO}(x) = \Ex_{q} \Big[ \log \frac{p(x,z)}{q(z)} \Big]$, the units of $p(x,z)$ and $q(z)$ do not cancel before the logarithm, and therefore the result is affected by changing our units of measurement from meters to feet.  However, the effect is to add a constant, and so this semantic point becomes irrelevant when the quantity is used as a loss function.

\section{Proofs}

First, some extra details for \cref{prop:logccave}.
By parameteriations $\mathbb P$ log-concave, we mean that, for every $a \in \Ar$, and $(s,t) \in \V(\Src a,\Tgt a)$, the function
$$
   \theta \mapsto -\log \p_a^{\,\theta}(\Tgt a =t\mid\Src a = a) \quad: \Theta_a \to [0,\infty]
$$
is convex.
This is true for many families of distributions of interest.
For example, if $\Src a, \Tgt a$ is discrete, and the cpd is parameterized
by stochastic matrices $\theta := \mat P = [p_{s,t}] \in [0,1]^{\V(\Src a, \Tgt a)}$, then
\[
   - \log \p_a^{\mat P}(\Tgt a =t | \Src a=s) = - \log (p_{s,t})
\]
which is clearly convex in $\mat P$.

To take another example: if $\p_a$ is linear Gaussian, \ie,
$\p_a(T|S) = \mathcal N(T | \mat A s + b,  \sigma^2)$, parameterized by
$(\mat A, b, 1/\sigma^2)$, then
\begin{align*}
   - \log \p_a^{(\mat A, b, \sigma^2)}(t|s)
   &= -\frac12 \log \frac{2\pi}{\sigma^2}  + \frac12 \left(\frac{t- \mat A s + b}{\sigma}\right)^2
\end{align*}
which is convex in $(\mat A, b, \frac1{\sigma^2})$.  Now, for the proof.

\recall{prop:logccave}
\begin{lproof} 
       \label{proof:logccave}
   By definition,
   \begin{align*}
       \aar*{\varphi \odot
             \dg M(\theta)}            
       &= \inf_\mu 
       \Big\{ 
           \gamma \SDef_{\dg M(\theta)}(\mu)
           + \OInc_{\dg M(\theta)}(\mu)  
       \Big\}.
   \end{align*}
   Only the final term actually depends on $\theta$, though. Let $F(\mu)$ capture
       the first term. For all of our examples, and indeed, if $\bbeta \ge \gamma\balpha$, it will be convex in $\mu$ \cite[from the proof of Proposition 3.2]{pdg-aaai}.
   Then we have
   \begin{align*}
       \aar*{\varphi \odot 
            \dg M}
           &= \inf_\mu  \left( F(\mu) + \Ex_\mu \left[ \sum_{\ed a ST}\beta_a \log \frac{\mu(T|S)}{\p_a(T|S)} \right] \right) \\
           &= \inf_\mu  \left( F(\mu) + \Ex_\mu \left[ \sum_{\ed a ST} \beta_a\log \frac{\mu(T|S)}{\lambda(T|S)} \right] +
           \Ex_\mu \left[\sum_{\ed a ST} \beta_a \log \frac{\lambda(T|S)}{\p_a(T|S)} \right] \right)
   \end{align*}
   The second term is then entropy (relative to the base distribution), which is
       convex in $\mu$. The first term, $F(\mu)$, is convex in $\mu$ as well, and neither depend on $\theta$. The final term is linear in $\mu$.
       Since $\mathbb P$ is log-concave in $\theta$, each $\log\frac{\lambda(t|s)}{\p_a(t|s)}$ for $a \in \Ar$ convex in $\theta$, and so that third term is a conic combination of terms that are all convex, and hence itself convex in $\theta$.
       Thus, the sum of all three terms in the infimum is jointly convex in $\theta$ and in $\mu$. Taking an infimum over $\mu$ pointwise, the result is still convex in $\theta$.
\end{lproof}

\recall{prop:LIR-EM}

\begin{lproof}\label{proof:LIR-EM}
Let $\dg M_{q,\theta}$ denote the PDG in the statement with observation $X=x$, generative arc labeled by $p_\theta$, and an inference arc $q$ into $Z$ with $\beta=\infty$ (high confidence). By \citet[Proposition 11]{one-true-loss} (``negative ELBO equals inconsistency''),
\[
\aar[\big]{\dg M_{q,\theta}} = -\mathrm{ELBO}_{p,q}(x)
= -\Ex_{q(z)}\Big[\log \tfrac{p(x,z;\theta)}{q(z)}\Big]
= \kldiv[\big]{q(z)}{p(z\mid x;\theta)}-\log p(x;\theta).
\]
Thus, for fixed $p(\cdot;\theta)$, minimizing $\aar[\big]{\dg M_{q,\theta}}$ over $q$ is equivalent to minimizing
$\kldiv[\big]{q}{p(z\mid x;\theta)}$.

Full control of $q$ (E-step):
Fix $\theta$. Since $\kldiv{\cdot}{\cdot} \ge 0$ with equality iff its two arguments agree,
\[
q^{*} \in \arg\min_{q} \aar[\big]{\dg M_{q,\theta}}
\quad\Longleftrightarrow\quad
q^{*}(z) = p(z\mid x;\theta^{(t)}_{LIR}).
\]
Therefore a LIR step that grants full control to $q$ drives $q$ to the exact posterior under the current $\theta$. This is precisely the \emph{E-step} of EM.

Full control of $p$ (M-step):
Now fix $q$ (in particular, after the E-step we have $q(z)=p(z\mid x;\theta^{(t)}_{LIR})$). Minimizing $\aar[\big]{\dg M_{q,\theta}}$ over $p$ (equivalently, maximizing the ELBO over $\theta$) is
\[
\theta^{(t+2)}_{LIR}
\in \arg\max_{\theta}
\Ex_{q(z)}\!\big[\log p(x,z;\theta)\big]
=
\arg\max_{\theta}
\Ex_{z\sim p(z\mid x;\theta^{(t)}_{LIR})}\!\big[\log p(x,z;\theta)\big],
\]
which is exactly the standard \emph{M-step} update.

Identification with EM iterates:
Let $\theta^{(0)}_{\text{LIR}}=\theta^{(0)}_{\text{EM}}$.
A LIR cycle consists of two refocus operations:
(i) full control of $q$ $\Rightarrow$ $q^{(t+1)}=p(z\mid x;\theta^{(t)}_{\text{LIR}})$ (E-step),
(ii) full control of $p$ $\Rightarrow$ $\theta^{(t+2)}_{\text{LIR}}\in\arg\max_{\theta}\Ex_{q^{(t+1)}}[\log p(x,z;\theta^{(t)}_{LIR})]$ (M-step).
The second update matches the EM M-step with posterior taken at $\theta^{(t)}_{\text{EM}}$.
Hence, by induction on $t$, $\theta^{(t)}_{\text{EM}} = \theta^{(2t)}_{\text{LIR}}$.

The optional {\sc Refocus} re-sampling of $x$ simply yields the usual (mini-)batched variant without affecting the maximizers of the conditional objectives. This proves that alternating full control between $q$ and $p$ in LIR implements EM as claimed.
\end{lproof}

\recall{prop:GAN}

\begin{lproof}\label{proof:GAN}
Let $C\!\in\!\{\real,\fake\}$ denote the coin (real vs.\ fake) and let $X$ be the displayed image. In the PDG in the text we have high-confidence arcs enforcing
\[
C \sim \mathrm{Bernoulli}\!\left(\tfrac12\right),\qquad
(X\mid C{=}\real) \sim \pdata,\quad
(X\mid C{=}\fake) \sim G,
\]
so the unique $\mu^*$ attaining the infimum in the observational inconsistency is the induced joint
\(
\mu^*(x,c)=\tfrac12\,\pdata(x)\,\mathbbm 1\{c{=}\real\}
+\tfrac12\,G(x)\,\mathbbm 1\{c{=}\fake\}.
\)
Write $D(c\mid x)$ for the discriminator cpd and $e(c\mid x)\equiv\tfrac12$ for the “equal coin” belief. The inconsistency of the PDG is obtained as follows.

\begin{align*}
&\aar{\dg M}
= \OInc_{\dg M}(\mu^\star)\\
&= \beta_D \Ex_{(X,C)\sim \mu^\star}\!\Big[\log \mu^\star(C\mid X) - \log D(C\mid X)\Big]
+ \beta_e \Ex_{(X,C)\sim \mu^\star}\!\Big[\log \mu^\star(C\mid X) - \log e(C\mid X)\Big] \\
&= -\beta_D \Ex_{(X,C)\sim \mu^\star}\!\big[\log D(C\mid X)\big]
   -\beta_e \Ex_{(X,C)\sim \mu^\star}\!\big[\log e(C\mid X)\big]
   + (\beta_D + \beta_e)\Ex_{(X,C)\sim \mu^\star}\!\big[\log \mu^\star(C\mid X)\big]\\
&= -\beta_D \Big(
        \tfrac12\,\Ex_{x\sim \pdata}\!\log D(x)
      + \tfrac12\,\Ex_{x\sim G}\!\log (1-D(x))
    \Big)
   -\beta_e \!\!\! \Ex_{(X,C)\sim \mu^\star\vphantom{\big|}}\!\!\!\!\!\big[\log e(C\mid X)\big]
   + (\beta_D + \beta_e)\!\!\!\!\Ex_{(X,C)\sim \mu^\star\vphantom{\big|}}\!\!\!\!\!\!\big[\log \mu^\star(C\mid X)\big]\\
&= -\beta_D \mathcal L^{\mathrm{GAN}}(G,D)
   -\beta_e \Ex_{(X,C)\sim \mu^\star}\!\big[\log e(C\mid X)\big]
   + (\beta_D + \beta_e)\Ex_{(X,C)\sim \mu^\star}\!\big[\log \mu^\star(C\mid X)\big]\\
&= -\beta_D \mathcal L^{\mathrm{GAN}}(G,D)
   -\beta_e \log \tfrac12
   + (\beta_D + \beta_e)\Ex_{(X,C)\sim \mu^\star}\!\big[\log \mu^\star(C\mid X)\big]\\
&= -\beta_D \mathcal L^{\mathrm{GAN}}(G,D)
   -\beta_e \log \tfrac12
   +(\beta_D + \beta_e) \Big(
        \tfrac12\,\Ex_{x\sim \pdata}\!\log \mu^\star(\real\mid x)
      + \tfrac12\,\Ex_{x\sim G}\!\log \mu^\star(\fake\mid x))
    \Big)\\
&= -\beta_D \mathcal L^{\mathrm{GAN}}(G,D)
   -\beta_e \log \tfrac12
   +(\beta_D + \beta_e) \Big(
        \tfrac12\,\Ex_{x\sim \pdata}\!\log \tfrac{\pdata(x)}{\pdata(x)+G(x)}
      + \tfrac12\,\Ex_{x\sim G}\!\log \tfrac{G(x)}{\pdata(x)+G(x)})
    \Big)\\
&= -\beta_D \mathcal L^{\mathrm{GAN}}(G,D)
   + \beta_e \log 2
   + (\beta_D + \beta_e)\Bigg(\Big(
      \tfrac12\,\Ex_{x\sim \pdata}\!\log \tfrac{\pdata(x)}{m_G(x)}
    + \tfrac12\,\Ex_{x\sim G}\!\log \tfrac{G(x)}{m_G(x)}
    \Big)
    - \log 2
    \Bigg)\\
&\qquad\qquad\text{(with }m_G(x)=\tfrac12\pdata(x)+\tfrac12 G(x)\text{)}\\
&= -\beta_D \mathcal L^{\mathrm{GAN}}(G,D)
   \;+\;(\beta_D+\beta_e)\mathrm{JS}(\pdata, G)
   \;-\;\beta_D \log 2,
\end{align*}

where $\mathrm{JS}(p,q)$ is the Jensen-Shannon divergence between $p$ and $q$.

\paragraph{Discriminator focus $(\varphi_D,\chi_D)$.}
By construction, $\chi_D$ gives control only over $D$, and $\varphi_D$ attends to the $D$ arc with weight $\beta_D=1$ while \emph{disbelieving} the $e$ arc, \ie, $\beta_e=-1$. The inconsistency simplifies to $\aar{\dg M} = -\mathcal L^{\mathrm{GAN}}(G,D) - \log 2$. Thus, minimizing inconsistency over $D$ is equivalent to the usual $\max_D \mathcal L^{\mathrm{GAN}}(G,D)$ step.

\paragraph{Generator focus $(\varphi_G,\chi_G)$.}
By construction, $\chi_G$ gives control only over $G$, and $\varphi_G$ \emph{disbelieves} the $D$ arc ($\beta_D=-1$) while attending to the $e$ arc with $\beta_e=1$. The inconsistency simplifies to $\aar{\dg M} = \mathcal L^{\mathrm{GAN}}(G,D) + \log 2$. Thus minimizing inconsistency over $G$ is equivalent to the usual $\min_G \mathcal L^{\mathrm{GAN}}(G,D)$ step.

Therefore, alternating LIR steps that refocus between $(\varphi_D,\chi_D)$ and $(\varphi_G,\chi_G)$ performs exactly the standard min-max optimization for GANs.
\end{lproof}

\recall{prop:bp}
\begin{lproof} \label{proof:bp}
When $\gamma=1$, and $\alpha, \beta = 1$ for all of the input factors, then the optimal
distribution $\mu^*$ that realizes the infimum is just the product of factors. It follows that any distribution that has those marginals will minimize the observational inconsistency.

Different interleavings of the update equations  \eqref{eq:X->a} and \eqref{eq:a->X} for different adjacent pairs $(a, X)$, which correspond to the different possible message passing schedules, also correspond immediately to different selections from this fixed set of foci in the LIR procedure.
\end{lproof}

\recall{prop:transformer}
\begin{lproof}\label{proof:transformer}
    Recall the parametric PDG in question: 
    \[
        \dg M(W_V, W_Q, W_K, x, x') :=
        \usebox\transformerpdg,
    \]
    where $\p_{ij}(X'_j \mid X_i) = \mathcal N(X'_j\mid W_V X_i, I)$ is a unit normal distribution.
    For completeness, and to aid the discussion, we will start with an analysis of the more general case where $\p_{ij}(X'_j \mid X_i) = \mathcal N(X'_j\mid W_V X_i, \Sigma_{ij})$ has an arbitrary covariance.
    
    Because of the deterministic arrows $x$ and $x'$, the infimum defining the inconsistency  must be attained at the degenerate distribution $\mu^*(X,X')$ where all mass is supported on the joint assignment $(x, x')$.

    We can therefore calculate
    \begin{align*}
        &\aar[\Big]{\dg M(W_Q, W_K, W_V, x, x')} \\
        &= \sum_{i,j \in [n]}
            \varphi_{ij} \kldiv[\Big]{\delta_{x'_j}(X'_j)}
                {\mathcal N(X'_j \mid W_{\!V} x_i, \, \Sigma_{ij})}
        \\&=
        \sum_{i,j \in [n]}
             - \varphi_{ij}  \log \Bigg[
                    (2\pi)^{-d/2} \det(\Sigma_{ij})^{-1/2}
                    \exp\Big(-\frac12 (x_j' - v_i)^{\sf T} \Sigma_{ij}^{-1}  (x_j' - v_i) \Big)
            \Bigg]  & \text{(see \cref{sec:density-note})}
        \\&=
        \sum_{i,j \in [n]} \frac{\varphi_{ij}}{2}
            \Big[ \log((2\pi)^d \det \Sigma_{ij}) + (x_j' - v_i)^{\sf T} \Sigma_{ij}^{-1}  (x_j' - v_i) \Big],
    \end{align*}
    where we have defined $v_i := W_V x_i$, as in the main text.

    What happens when we control $x'$? 

    The equation above is (strongly) convex in $x'$, and so gradient descent moves to the unique local minimum.
    The directional derivative of $\aar{\dg M}$ with respect to $x'$ in direction $u = (u_j)_{j=1}^n$ is given by
    \begin{align*}
        &\frac{\mathrm d}{\mathrm d\epsilon} 
         \sum_{i,j \in [n]} \frac{\varphi_{ij}}{2}
           (u_j \epsilon + x_j' - v_i)^{\sf T} \Sigma_{ij}^{-1}  (u_j \epsilon + x_j' - v_i)
           ~\bigg|_{\epsilon=0} 
        \\
        &= \sum_{i,j \in [n]} \frac{\varphi_{ij}}{2}
        \bigg(
            \frac{\mathrm d}{\mathrm d\epsilon} \Big[\epsilon^2 u_j^{\sf T} \Sigma_{ij}^{-1} u_j\Big]_{\epsilon{=}0}
            + 
            \frac{\mathrm d}{\mathrm d\epsilon} \Big[\epsilon
                u_j^{\sf T} \Sigma_{ij}^{-1} (x_j' - v_i) + \epsilon (x'_j - v_i)^{\sf T} \Sigma_{ij}^{-1} u_j
            \Big]_{\epsilon{=}0}
            \\&\hspace{2in}
            +
            \frac{\mathrm d}{\mathrm d\epsilon} \Big[(x_j' - v_i)^{\sf T} \Sigma_{ij}^{-1}  (x_j' - v_i)\Big]
        \bigg)
        \\&=
        \sum_{i,j \in [n]}  \frac{\varphi_{ij}}{2} u_j^{\sf T} (\Sigma_{ij}^{-1} + \Sigma_{ij}^{-\sf T})(x'_j - v_i)
        \\&=
        \sum_{i,j \in [n]} \varphi_{ij} u_j^{\sf T} \Sigma_{ij}^{-1}(x'_j - v_i)
            & \hspace{-2in}\text{since $\Sigma^{-1}_{ij}$ is symmetric}
        \\&=
        \sum_{j \in [n]} u_j^{\sf T} \sum_{i \in [n]} \varphi_{ij} \Sigma_{ij}^{-1} (x_j' - v_i)
        .
    \end{align*}
    The unique attractor $x'_*$ of the ODE defined by these gradients 
    is characterized by having zero directional gradients for all $\vec u$, 
    which is the case iff
    \begin{align*}
        &\forall j.~ \sum_i \varphi_{ij} \Sigma_{ij}^{-1} (x_j' - v_i) = 0
        \\\iff\qquad&
        \forall j.~\Big(\sum_{i\in [n]} \varphi_{ij} \Sigma_{ij}^{-1} \Big) x_j' = \sum_{i\in [n]} \varphi_{ij} \Sigma_{ij}^{-1} v_i
        \\\iff\qquad&
        \forall j.~x'_j =  \Big(\sum_{i\in [n]} \varphi_{ij} \Sigma_{ij}^{-1} \Big)^{-1}\sum_{i\in [n]} \varphi_{ij} \Sigma_{ij}^{-1} v_i.
            &\text{provided the matrix on the left is invertible.}
    \end{align*}
    In the special case where $\Sigma_{ij} = \sigma_{ij}^2 I$,
        and let $\psi_{ij} := {\varphi_{ij}}/{\sigma^2_{ij}}$,
     the optimum is 
    \[
        x'_j = \sum_{i\in [n]} \Big(\frac{\psi_{ij}}{\sum_{i' =1}^n \psi_{i'j}}\Big) v_i.
    \]
    Therefore, 
    whenever it is the case that
    $\psi_{ij} = b \exp \langle k_i, q_j\rangle$ for some constant $b$, 
    it is also the case that the unique fixed point of LIR when controlling $x'$ leads to $x'_j = \sum_i \alpha_{i|j} v_i$ where $\alpha_{i|j} = \mathrm{softmax}_i [\langle k_i, q_j \rangle]_{i,j} 
    \displaystyle=\Big(\frac{\exp \langle k_i, q_j \rangle }{\sum_{i' =1}^n \exp \langle k_{i'}, q_j \rangle }\Big)
    $. 
    
    In particular, this is the case when $\varphi_{ij} = \exp \langle k_i, q_j\rangle$ and $\sigma_{ij} = 1$, as assumed in the theorem statement.
\end{lproof}

\recall{prop:GFN}
\begin{lproof}\label{proof:GFN}
    Recall the PDG in question:
    \[
        \dg M(\theta) = \;\usebox\gfnpdg .
    \]

    Let $\hat Q(\tau,i) := Q(\tau)\,\mathrm{Unif}(i\mid \tau)$ denote the distribution over trajectories and indices obtained by first drawing a trajectory $\tau\sim Q$ and then a uniformly random index $i\in\{0,\ldots,n\}$ along the (random) trajectory $\tau=(s_0,\ldots,s_n{=}x,s_{n+1}{=}s_{\!f})$. Because the labels on $Q(\tau)$ and $\mathrm{Unif}(i\mid \tau)$ are high confidence, the infimum in the observational inconsistency is attained by the induced joint
    \[
    \mu^*(\tau,i,S,S')
    \;=\;
    Q(\tau)\,\mathrm{Unif}(i\mid \tau)\;
    \delta(S=\tau[i])\;
    \delta(S'=\tau[i{+}1]).
    \]

    \paragraph{Subclaim.}
    For all $s\in\mathcal S\setminus\{s_0,s_{\!f}\}$ we have $\mu^*(S{=}s)=\mu^*(S'{=}s)$; also, $\mu^*(S{=}s_0)=\mu^*(S'{=}s_{\!f})$.

    \emph{Proof.}
    Since the zeroth element $\tau[0]$ of a trajectory $\tau$ is always $s_0$, the assumption that $S = \tau[i] = s \notin \{ s_0, s_{\!f} \}$ necessitates $i > 0$. 
    Therefore,
    \begin{align*}
        \mu^*(S=s) 
            &= \hat Q(\{ (\tau,i) : \tau[i] = s\}) \\
            &= \hat Q(\{ (\tau,i-1) : \tau[i] = s\})  
                &\text{ by uniformity of $i$, as $i > 0$} \\
            &= \hat Q(\{(\tau, j) : \tau[j+1] = s\}) 
                &\text{ by taking $j:=i-1$ } \\
            &= \mu^*(S' = s).
    \end{align*}
    Summing over $s$ and using $\mu^*(S{=}s_{\!f})=\mu^*(S'{=}s_0)=0$ yields $\mu^*(S{=}s_0)=\mu^*(S'{=}s_{\!f})$. \qed

    \medskip
    Only the two cpds $P_F(\cdot\mid\cdot)$ and $P_B(\cdot\mid\cdot)$ contribute non-constant terms (all copy/selection hyperarcs are satisfied by $\mu^*$). Denote
    \[
      a(\tau) := \log P_F(\tau)=\sum_{i=0}^{n}\log P_F(s_{i+1}\mid s_i),
      \qquad
      b(\tau) := \log P_B(\tau)=\sum_{i=0}^{n}\log P_B(s_i\mid s_{i+1}).
    \]
    By construction at the sink,
    \[
      P_B(\tau)
      \;=\; P_B(\tau\mid x)\,\frac{R(x)}{Z}
      \quad\Longrightarrow\quad
      b(\tau)=\log P_B(\tau\mid x)+\log R(x)-\log Z.
    \]

    The observational incompatibility (with centered surprisal attention) is
    \begin{align*}
      \OInc_{\dg M}(\mu^*)
      \;=\;
      \Ex_{(\tau,i)\sim\hat Q}
      \Big[&
        \overbrace{\varphi(P_F)}^{=\,b(\tau)-a(\tau)}
          \!\cdot\,
          \overbrace{\kldiv{\delta_{S'=\tau[i+1]}}{P_F(\,\cdot\mid S=\tau[i])}}^{\,=\;-\log P_F(\tau[i{+}1]\mid\tau[i])}
        \\[-2ex]&
        +\;\underbrace{\varphi(P_B)}_{=\,a(\tau)-b(\tau)}
          \!\cdot\,
          \underbrace{\kldiv{\delta_{S=\tau[i]}}{P_B(\,\cdot\mid S'=\tau[i+1])}}_{\,=\;-\log P_B(\tau[i]\mid\tau[i{+}1])}
      \;\Big]
      .
    \end{align*}
    Since the attention $\varphi$ depends only on $\tau$, we can factor it out of the expectation over $i$, which is uniform over the $|\tau|$ transition sin the trajectory, and expectation amounts to their sum divided by $|\tau|$:
    just sums the local logs back to the whole-trajectory logs:
    \[
      \OInc_{\dg M}(\mu^*)
      \;=\;
      \Ex_{\tau\sim Q}
      \Big[
        \big(b(\tau)-a(\tau)\big)\,\big(\frac{-a(\tau)}{|\tau|}\big)
        \;+\;
        \big(a(\tau)-b(\tau)\big)\,\big(\frac{-b(\tau)}{|\tau|}\big)
      \Big]
      .
    \]
    Expanding,
    \[
      \OInc_{\dg M}(\mu^*)
      \;=\;
      \Ex_{\tau\sim Q}\!\Big[ \frac{ a(\tau)^2 + b(\tau)^2 - 2\,a(\tau)b(\tau) }{|\tau|}\Big] 
      \;=\;
      \Ex_{\tau\sim Q}\!\Big[ \frac{1}{|\tau|} \big(a(\tau)-b(\tau)\big)^2 \Big] 
      .
    \]
    Using the sink identity for $b(\tau)$, we obtain
    \[
      a(\tau)-b(\tau)
      \;=\;
      \log\frac{P_F(\tau)\,Z}{R(x)\,P_B(\tau\mid x)}.
    \]
    Therefore
    \[
      \OInc_{\dg M}(\mu^*)
      \;=\;
      \Ex_{\tau\sim Q}\!
      \Bigg[\,
        \frac1{|\tau|}
        \log^{\!2}\!
        \frac{P_F(\tau)\,Z}{R(x)\,P_B(\tau\mid x)}
      \Bigg]
      \;=\;
      \mathcal L_{\mathrm{ModTB}}(Q)
      ,
    \]
    which proves that the (centered-surprisal) attention yields inconsistency equal to the modified TB objective.

    \paragraph{Gradients.}
    Treat $\varphi$ as a fixed (stop-gradient) attention. If only $P_F$ depends on $\theta$,
    \[
      \nabla_\theta \OInc_{\dg M}
      \;=\;
      2\,\Ex_{\tau\sim Q}
      \Bigg[
        \log\frac{P_F(\tau)\,Z}{R(x)\,P_B(\tau\mid x)}\;\nabla_\theta \log P_F(\tau)
      \Bigg],
    \]
    which is exactly $\nabla_\theta \mathcal L_{\mathrm{TB}}(Q)$. If, in addition, $P_B$ shares the same parameters (or we also place control on $P_B$), an equal-magnitude term appears with opposite sign in $\nabla_\theta \log P_B(\tau\mid x)$, yielding the same direction and a proportionality factor (a constant multiple), as claimed. Hence LIR on $\dg M$ with the stated attention and control over $P_F$ (and optionally $P_B$) trains a GFlowNet with the TB loss.
\end{lproof}

\recall{prop:eumaxmax}
\begin{lproof}\label{proof:eumaxmax}
   Since the choice $a$ is deterministic, the value of $A$ is determined; likewise the value of $\Tru$ is also fixed. 
   Similarly, as $u$ is a deterministic function, the value of $U$ is determined according to $u$.
   Thus we need only consider distributions $\mu(S, O)$ in our minimization; the other variables can be found as a function of these.
   However, we also know that $\mu(O\mid S) = \tau(O|S,A{=}a)$ since it is given with high confidence. 
   Therefore it suffices to restrict our search to distributions over the variable $S$.
   
   To simplify notation, let 
   $ EU(s,a) := \Ex_{o \sim \tau (O|s,a)} [ u(o) ] + \log k$
   be the expected utility of an action, shifted by the constant $\log k$.
   Note that
   \[
       \log \frac{1}{b(\Tru{=}\trut\mid U{=}u(o))}
       = - \log (k \cdot \exp( u(o) ))
       = - u(o) - \log k,
   \]
   which in expectation over $\tau(O|s,a)$, is $ - EU(s,a)$. 
   With this in mind, we calculate:
   \begin{align*}
       \aar[\big]{\dg M_{p,\tau,u,b}+A{=}a}_{\gamma}
       &= \inf_{\mu(S)} 
           \beta_p \Ex_{s \sim \mu} \Big[ \log \frac {\mu(s)}{p(s)} 
           + \frac{\beta_b}{\beta_p} \Ex_{o\sim \tau|a,s}\Big[\log \frac1{b(\Tru{=}\trut\mid U{=}u(o))}\Big]
           \\
       &= \inf_{\mu(S)} 
           \beta_p \Ex_{s \sim \mu} \Big[ \log \frac {\mu(s)}{p(s)} 
           + \log \circ \exp \Big(- \frac{\beta_b}{\beta_p} EU(s,a) \Big)
           \\
       &= \inf_{\mu(S)} 
           \beta_p \Ex_{s \sim \mu} \Big[ \log \frac {\mu(s)}{1}\cdot\frac{\exp(-\frac{\beta_b}{\beta_p} EU(s,a))}{p(s)} \cdot \frac ZZ \Big]
           \\
       &= \inf_{\mu(S)} 
           \beta_p \Ex_{s \sim \mu} \Big[ \log \frac {\mu(s)}{1}\cdot\frac{Z}{p(s)\cdot \exp(\frac{\beta_b}{\beta_p} EU(s,a))} \cdot \frac 1Z \Big]
               \numberthis\label{eq:save-eu-calc}
   \end{align*}
   At this point, we can use the same trick used repeatedly in the proving results of \citet{one-true-loss}: take $Z$ to be the normalization constant needed to regard the middle fraction as the inverse of a probability distribution $\nu$. 
   Once we do so, we are left with an infimum over a KL divergence $\kldiv{\mu}{\nu}$ plus the expectation of a constant:
   \begin{align*}
       \aar[\big]{\dg M_{p,\tau,u,b}+A{=}a}_{\gamma}
           &= \beta_p \log \frac1 Z
           = - \log \sum_{s} p(s) \exp\Big( + \frac{\beta_b}{\beta_p} EU(s,a)\Big).
   \end{align*}
   We now look at the two extreme cases. 
   
   When $\beta_p \to \infty$, then
       the ratio $\frac{\beta_b}{\beta_p}$ becomes small, 
       and we can use the fact that $\exp( \epsilon) \approx 1+\epsilon$
       for small $\epsilon$
   to find that the inconsistency of interest is
   approximately 
   \begin{align*}
       \approx 
       - \beta_p \log \sum_{s} p(s) \Big[ 1 + \frac{\beta_b}{\beta_p} EU(s,a) \Big]
       = - \beta_p \log( 1 +  \frac{\beta_b}{\beta_p} EU(s,a) )
       \approx
       - \beta_b EU(s,a)
   \end{align*}
   Alternatively, more directly, when $\beta_p = \infty$, the optimal distribution must be $\mu = p$, and so the inconsistency is immediately $ - \Ex_{s \sim p} [ \beta_b EU(s,a)]$. 
   Either way, minimizing this quantity amounts to maximizing expected utility, as the two differ by a negative affine transformation.
   
   At the other extreme, when $\beta_b \to \infty$, 
   we can write our expression in terms of 
   $\mathrm{LSE} \{ x_1, \ldots, x_n \} := \log \sum_{i=1}^n \exp( x_i)$ (LogSumExp)
   which is a smooth approximation to a $\max$.  (In a moment, we will negate its arguments and the final output, using it as an approximation to a $\min$, instead.)
   Picking back up from \eqref{eq:save-eu-calc} and letting $t := \frac{\beta_b}{\beta_p}$, we find
   \begin{align*}
       \aar[\big]{\dg M_{p,\tau,u,b}+A{=}a}_{\gamma}
           &= - \beta_b \cdot \frac{-1}{t} \mathop{\mathrm{LSE}}\limits_{s \in \V S} \Big[-t \cdot \Big( \frac1t \log \frac1{p(s)} -  EU(s,a)\Big) \Big]. 
   \end{align*}
   Using the standard fact
   \unskip\footnote{
       Letting $m := \min_i x_i$, observe that, for all $t > 0$, we have
       $\exp(-tm) \le \sum_i \exp(-t x_i) \le n \exp(- tm)$.
       Apply a logarithm and multiply by $-\frac{1}{t}$ to get the promised result.}
   that 
   \[
       \min_{i \in [n]} x_i - \frac{1}{t} \log n \le 
           \frac{-1}{t} \mathop{\mathrm{LSE}}\limits_{i \in [n]} ( - t x_i ) < \min_{i \in [n]} x_i,
   \]
   we find that, in our case,
   \begin{equation}
   M - \frac{1}{t} \log |\V S|
   ~~\le~~ \frac{1}{\beta_b} \aar[\big]{\dg M_{p,\tau,u,b}+A{=}a}_{\gamma}
   ~~\le~~ M
   \end{equation}
   where $M := \min_{s \in \V S}  ( -  EU(s,a)  + \frac{1}{t} \log \frac 1{p(s)} )$.
   In particular, when $\beta_b \to \infty$, meaning $t \to \infty$, the gap between the upper and lower bounds shrinks to zero,
   and the resulting inconsistency becomes
   $ - \min_{s \in \V S} ( - EU(s,a) ) = \max_{s \in \V S)} EU(s,a)$,
   proving the result.
\end{lproof}

\section{Generative Flow Network Experiments}
\label{apx:gfn}

\subsection{Trajectory-Length Normalization for the Trajectory Balance and Log-Partition Variance Losses}

In Trajectory Balance (TB) and Log-Partition Variance (LPV, also called VarGrad) objectives, longer trajectories accumulate larger log-probability sums, causing per-trajectory squared losses to scale with trajectory length. This can starve gradients on shorter trajectories and bias learning toward long-trajectory corrections. Their losses (reproduced below) were found to be missing a normalization constant $n_i$, denoting the length of the trajectory $\tau_i$.

The standard trajectory balance (TB) condition states that for all complete
trajectories $\tau = (s_0 \to s_1 \to \cdots \to x)$, $Z \prod_{t} P_F(s_{t+1} \mid s_t) =  R(x) \prod_{t} P_B(s_t \mid s_{t+1})$, or equivalently,

\[
  \log Z + \log P_F(\tau)
  \;=\;
  \log R(x) + \log P_B(\tau \mid x).
\]

which gives the loss 

\[
   \mathcal{L}_\mathrm{TB}(Q) = \Ex_{\tau \sim Q}\left[ \log^2 
         \frac{P_F(\tau) Z}{R(x) P_B(\tau\mid x)}  
      \right].
\]

When trained as a GFlowNet, the forward policy $P_F$ and log partition function estimate $\log Z$ are parameterized separately, optionally alongside the backward policy $P_B$ (which may not be parameterized as it is possible to leave it fixed during training), and $R(x)$ is a reward function providing some scalar reward defining a distribution over the state space $\mathcal{X}$.
 
In contrast, the Log Partition Variance loss avoids explicit estimation of $\log Z$ by noting that it is constant with respect to~$\tau$. Given the per-trajectory residual,

\[
  \Delta(\tau)
  \;=\;
  \log P_F(\tau) - \log R(x) - \log P_B(\tau \mid x).
\]
$\Delta(\tau) = -\log Z$ when the TB condition holds exactly, so $\log Z$ can treated as a constant shift and dropped from the variance:
\[
    \mathcal{L}_{\mathrm{LPV}}(Q)
    \;=\;
    \mathrm{Var}_{\tau \sim Q}\!\left[
      \log \frac{P_F(\tau)}{R(x)\, P_B(\tau \mid x)}
    \right]  
\]
Expanding the variance via $\mathrm{Var}[X] = \Ex[X^2] - (\Ex[X])^2$:
\[
  \mathcal{L}_{\mathrm{LPV}}(Q)
  \;=\;
  \Ex_{\tau \sim Q}\!\left[
    \left(
      \log \frac{P_F(\tau)}{R(x)\, P_B(\tau \mid x)}
    \right)^{\!2}
  \right]
  \;-\;
  \left(
    \Ex_{\tau \sim Q}\!\left[
      \log \frac{P_F(\tau)}{R(x)\, P_B(\tau \mid x)}
    \right]
  \right)^{\!2}.
\]

In practice, all expectations are estimated by the sample mean over a minibatch $\{\tau_i\}_{i=1}^m$ of trajectories sampled from the forward policy $P_F$. We normalize the per-trajectory squared loss by dividing by per-trajectory length $n_i$ after squaring, preserving the fixed point ($s_i + \log Z = 0$) and preventing an attenuation of gradients that would arise from normalizing the score before squaring.
For TB,  the modified loss is simply:

\begin{equation}
   \mathcal{L}_{\mathrm{ModTB}} := \frac{1}{m}\sum_{i=1}^{m} \frac{s_i^2}{n_i}, 
   \qquad \text{where}~~s_i = \log \frac{P_F(\tau_i)\, Z}{R(x_i)\, P_B(\tau_i \mid x_i)},
\end{equation}

and for LogPV, the modification is analogous. Let $s_i = \log \frac{P_F(\tau_i)}{R(x_i)\, P_B(\tau_i \mid x_i)}$ for trajectories $\tau_i \sim Q$, 

\begin{equation}
\mathcal{L}_{\text{ModLPV}} = \frac{1}{m}\sum_{i=1}^{m} \frac{(s_i - \bar{s})^2}{n_i},
\quad\text{where}\quad \bar{s} = \frac{1}{m}\sum_{i=1}^{n} s_i 
\quad\text{and}\quad
s_i = \log \frac{P_F(\tau_i)}{R(x_i)\, P_B(\tau_i \mid x_i)}
\end{equation}

\noindent where $T_i = |\tau_i|$ is the length of trajectory $i$. This is no longer
$\mathrm{Var}_Q[s]$; the $\frac{1}{T_i}$ weighting downweights long trajectories, analogous to SubTB-style 
per-step normalization \cite{madan2023learning}.

\subsection{Experimental Setup}

\subsubsection{Environment} 

All losses were tested on four variants of the discrete HyperGrid task: a $d$-dimensional grid of height $H$ where trajectories can terminate in any state and all trajectories start at $\mathbf{0}_d$. For all experiments, we evaluate on a 4-dimensional HyperGrid with height $H = 24$, giving a state space of $24^4 = 331{,}776$ terminal states (states are vectors $s \in \{0, 1, \ldots, 23\}^4$). We define a normalized coordinate $a_i = |s_i / (H - 1) - 0.5|$ for each dimension $i$.

\paragraph{Original:} The original hypergrid environments follows \cite{bengio2021flow}:

$$R(s) = R_0 + R_1 \prod_i \mathbf{1}[a_i > 0.25] + R_2 \prod_i \mathbf{1}[0.3 < a_i < 0.4]$$

with $R_0 = 0.1$, $R_1 = 0.5$, $R_2 = 2.0$. Intuitively, modes occupy the 4 corners of each dimension, giving $4^4 = \mathbf{256}$ mode states. Modes are states satisfying the inner band condition in all dimensions simultaneously (reward $\geq R_0 + R_1 + R_2 = 2.6$). At $H = 24$, the band indices per dimension are $\{3, 4, 19, 20\}$.

\paragraph{Cosine:} A smooth, oscillatory reward with per-dimension factors:

$$R(s) = R_0 + R_1 \prod_i f(a_i), \quad f(a) = (\cos(50a) + 1) \cdot \mathcal{N}(0,1)(5a)$$

where $\mathcal{N}(0,1)(\cdot)$ is the standard normal PDF. With $R_0 = 0.1$, $R_1 = 0.5$, and closeness parameter $\gamma = 0.8$, mode states are those with reward $\geq R_0 + (\gamma \cdot f_{\max})^4 \cdot R_1$, where $f_{\max} \approx 0.581$ is the discrete per-dimension maximum at $H = 24$. The product structure yields $\mathbf{1{,}280}$ mode states with a regular pattern across dimensions.

\paragraph{Bitwise XOR:} A tiered compositional reward based on GF(2) parity constraints:

$$R(s) = \sum_{t=0}^{2} w_t \cdot \mathbf{1}\!\left[\forall\, b \in [l_t, h_t],\; \bigoplus_{i=0}^{3} \mathrm{bit}_b(s_i) = 0\right]$$

with tier weights $w = (1, 10, 100)$ and bit ranges $(l_t, h_t) \in \{(0, 5), (0, 7), (0, 9)\}$. Each tier requires even parity (XOR $= 0$) across all 4 dimensions at every bit plane in its range. Rewards are cumulative: tier $t$ is awarded only if all constraints through tier $t$ are satisfied. Modes achieve the highest tier (reward $= 111$). Since values $\leq 23$ have at most 5 nonzero bits, the higher bit-plane constraints are automatically satisfied, and mode membership reduces to $s_0 \oplus s_1 \oplus s_2 \oplus s_3 = 0$ (bitwise). The fourth coordinate is determined ($s_3 = s_0 \oplus s_1 \oplus s_2$) subject to $s_3 < 24$, giving $\mathbf{10{,}752}$ mode states.

\paragraph{Multiplicative Coprime:} A tiered reward based on prime factorization constraints:

$$R(s) = \sum_{t=0}^{2} w_t \cdot \mathbf{1}\!\left[\forall\, i,\; s_i = \prod_{p \in \mathcal{P}} p^{e_{i,p}} \text{ with } e_{i,p} \leq c_t\right]$$

with $w = (1, 10, 100)$, prime set $\mathcal{P} = \{2, 3, 5\}$, and uniform exponent cap $c_t = 2$ for all tiers. Each dimension's value must factor completely over $\mathcal{P}$ with exponents $\leq 2$ (zero values are excluded; $s_i = 1$ is always valid). The valid values per dimension are $\{1, 2, 3, 4, 5, 6, 9, 10, 12, 15, 18, 20\}$ (12 values), yielding $12^4 = \mathbf{20{,}736}$ mode states.

\subsection{Training}

\paragraph{Hyperparameter Tuning:} To fairly compare each loss, we first used the Optuna \cite{optuna} TPE sampler to tune:

\begin{itemize}[nosep,topsep=.2ex,itemsep=.2ex]
    \item Learning rate: $[10^{-5}, 10^{-1}]$ (log-scale)
    \item Adam $\beta_2 \in \{0.99, 0.999, 0.9999\}$
    \item Gradient clipping norm: $[0.01, 10.0]$ (log-scale)
    \item LogZ learning-rate multiplier (TB variants only): $[10, 1000]$ (log-scale)
\end{itemize}

The LogZ learning-rate multiplier is commonly used to assist in the convergence of TB-based algorithms which cannot converge until the $log Z$ estimate is close to the true partition function, which may be very large for some distributions. To accomplish this, practitioners often train $log Z$ with an independent optimizer and/or learning rate. In our case the multiplier is applied to whatever base learning rate is proposed by Optuna for that run. For all experiments, batch size=1024, we employed a replay buffer with capacity=10,000 and 50\% replay fraction, a per-trajectory loss clamp of 100, linear LR schedule down to 1\% of the initial learning rate, and the AdamW optimizer with weight decay of 0.01. We ran 50 trials for each of the 16 algorithm-environment pairs, each run for 2,000 iterations with a single seed, optimizing final L1 distance estimated between the true reward distribution and the learned posterior using $10^6$ samples from the trained $P_F$. 

\paragraph{Model Training:} The top thee hyperparameter configurations found were brought forward, and trained across 5 seeds for 4,000 iterations each (240 runs total). The following metrics were collected:

\begin{itemize}[nosep,topsep=.2ex,itemsep=.2ex]
    \item \textbf{L1 distance} between the learned terminal-state distribution and the true reward-proportional distribution: $\sum_s |p(s) - p^*(s)|$
    \item \textbf{Jensen--Shannon divergence (JSD)}: $\frac{1}{2}\mathrm{KL}(p \| m) + \frac{1}{2}\mathrm{KL}(p^* \| m)$, where $m = \frac{1}{2}(p + p^*)$
    \item \textbf{Mode coverage}: fraction of reward modes discovered during training
\end{itemize}

\subsection{Results}

See Table \ref{apx:gfnrank1} for the median $\pm$ interquartile range metrics for all losses on all environments using the single best hyperparameters identified across 5 seeds. All models (with some single-seed exceptions) cover all environment modes. However, two methods that have 100\% mode coverage may have substantially different L1 / JSD distances. L1 \& JSD measures are highly correlated ($r>0.95\%)$. On the Bitwise XOR environment, the modified losses show clear benefits as measured using the distribution distance measures. ModLPV also shows faster convergence at the beginning of training relative to the LPV variant, as measured by the fraction of iterations required to reach $2\times$ the final loss. Differences appear to be environment dependent, and the advantages of the improved training dynamics of the modified losses may be even more apparent on more sophisticated and challenging reward distributions.

\begin{table}[h]
\centering
\caption{Final metrics for rank-1 configurations (median $\pm$ IQR across 5 seeds).}
\begin{tabular}{lcccc}
\toprule
Environment & \textbf{TB} & \textbf{ModTB} & \textbf{LPV} & \textbf{ModLPV} \\
\midrule
\multicolumn{5}{c}{\textit{Final L1 Distance}} \\
\midrule
Bitwise XOR & $1.74 \pm 0.28$ & \textbf{$1.23 \pm 0.01$} & $1.49 \pm 0.51$ & $1.24 \pm 0.00$ \\
Cosine & $0.449 \pm 0.000$ & $0.448 \pm 0.000$ & \textbf{$0.448 \pm 0.000$} & $0.448 \pm 0.000$ \\
Multiplicative Coprime & $0.137 \pm 0.003$ & $0.137 \pm 0.003$ & \textbf{$0.134 \pm 0.001$} & $0.138 \pm 0.002$ \\
Original & \textbf{$0.438 \pm 0.000$} & $0.438 \pm 0.000$ & $0.44 \pm 1.54$ & $0.439 \pm 0.001$ \\
\midrule
\multicolumn{5}{c}{\textit{Final JSD}} \\
\midrule
Bitwise XOR & $0.487 \pm 0.111$ & $0.289 \pm 0.006$ & $0.388 \pm 0.182$ & \textbf{$0.288 \pm 0.001$} \\
Cosine & $0.050 \pm 0.000$ & $0.050 \pm 0.000$ & \textbf{$0.050 \pm 0.000$} & $0.050 \pm 0.000$ \\
Multiplicative Coprime & $0.010 \pm 0.001$ & $0.010 \pm 0.000$ & \textbf{$0.008 \pm 0.000$} & $0.010 \pm 0.001$ \\
Original & \textbf{$0.052 \pm 0.000$} & $0.052 \pm 0.000$ & $0.052 \pm 0.620$ & $0.052 \pm 0.000$ \\
\midrule
\multicolumn{5}{c}{\textit{Mode Coverage (\%)}} \\
\midrule
Bitwise XOR & \textbf{$100.00 \pm 0.00$} & $99.27 \pm 0.06$ & $100.00 \pm 0.00$ & $99.45 \pm 0.29$ \\
Cosine & \textbf{$100.00 \pm 0.00$} & $100.00 \pm 0.00$ & $100.00 \pm 0.00$ & $100.00 \pm 0.00$ \\
Multiplicative Coprime & \textbf{$100.00 \pm 0.00$} & $100.00 \pm 0.00$ & $100.00 \pm 0.00$ & $100.00 \pm 0.00$ \\
Original & \textbf{$100.00 \pm 0.00$} & $100.00 \pm 0.00$ & $100.00 \pm 96.09$ & $100.00 \pm 0.00$ \\
\midrule
\multicolumn{5}{c}{\textit{Iterations to $2\times$ final loss (fraction, $\downarrow$)}} \\
\midrule
Bitwise XOR & $0.006 \pm 0.003$ & $0.030 \pm 0.023$ & \textbf{$0.004 \pm 0.002$} & $0.045 \pm 0.009$ \\
Cosine & $0.835 \pm 0.054$ & $0.857 \pm 0.058$ & $0.794 \pm 0.011$ & \textbf{$0.521 \pm 0.055$} \\
Multiplicative Coprime & \textbf{$0.199 \pm 0.356$} & $0.455 \pm 0.099$ & $0.574 \pm 0.044$ & $0.248 \pm 0.109$ \\
Original & $0.838 \pm 0.007$ & $0.895 \pm 0.013$ & $0.857 \pm 0.858$ & \textbf{$0.586 \pm 0.023$} \\
\bottomrule
\end{tabular}
\label{apx:gfnrank1}
\end{table}

\end{document}